\documentclass[final,3p,times]{elsarticle}

\usepackage{lineno,hyperref}
\usepackage{amssymb}

\usepackage{graphicx}
\usepackage{comment}
\usepackage{amsmath}
\usepackage{algorithm}
\usepackage[noend]{algpseudocode}
\usepackage{algorithmicx}
\usepackage{multirow}
\usepackage{soul}
\usepackage{color,xcolor}
\usepackage{subfigure} 

\modulolinenumbers[5]

\journal{Journal of \LaTeX\ Templates}

\begin{document}

\begin{frontmatter}

\title{A multi-stage semi-supervised improved deep embedded clustering method for bearing fault diagnosis under the situation of insufficient labeled samples}

\tnotetext[mytitlenote]{Fully documented templates are available in the elsarticle package on \href{http://www.ctan.org/tex-archive/macros/latex/contrib/elsarticle}{CTAN}.}


\author[mymainaddress]{Tongda Sun}
\author[mymainaddress]{Gang Yu\corref{mycorrespondingauthor}}
\ead{gangyu@hit.edu.cn}
\cortext[mycorrespondingauthor]{Corresponding author}
\address[mymainaddress]{School of Mechanical Engineering and Automation, Harbin Institute of Technology, Shenzhen, Shenzhen, Guangdong 518055, P.R. China}

\soulregister{\cite}7 
\soulregister{\citep}7 
\soulregister{\citet}7 
\soulregister{\ref}7 
\soulregister{\pageref}7 

\sethlcolor{yellow}

\begin{abstract}
{\color{black}Although data-driven fault diagnosis methods have been widely applied, massive labeled data are required for model training. However, a difficulty of implementing this in real industries hinders the application of these methods. Hence, an effective diagnostic approach that can work well in such situation is urgently needed.} {\color{black}In this study}, {\color{black}a multi-stage semi-supervised improved deep embedded clustering (MS-SSIDEC) method, which combines semi-supervised learning with improved deep embedded clustering (IDEC), is proposed to jointly explore scarce labeled data and massive unlabeled data.} In the first stage, {\color{black}a skip-connection-based convolutional auto-encoder (SCCAE) that can automatically map the unlabeled data into a low-dimensional feature space is proposed and pre-trained to be a fault feature extractor.} In the second stage, {\color{black}a semi-supervised improved deep embedded clustering (SSIDEC) network is proposed for clustering. It is first initialized with available labeled data and then used to simultaneously optimize the clustering label assignment and make the feature space to be more clustering-friendly. To tackle the phenomenon of overfitting, virtual adversarial training (VAT) is introduced as a regularization term in this stage.} In the third stage, pseudo labels are obtained by the high-quality results of {\color{black}SSIDEC}. The labeled dataset can be augmented by these pseudo-labeled data and then leveraged to train a bearing fault diagnosis model. {\color{black}Two public datasets of vibration data from rolling bearings are used to evaluate the performance of the proposed method. Experimental results indicate that the proposed method achieves a promising performance in both semi-supervised and unsupervised fault diagnosis tasks.} This method provides a new approach for fault diagnosis under the situation of limited labeled samples by effectively exploring unsupervised data.
\end{abstract}


\end{frontmatter}

\sethlcolor{yellow}

\section{Introduction}
{\color{black}Rolling bearings are} a key part of the rotating machinery. Because {\color{black}operating under} complex conditions of heavy load and high speed for a long time, {\color{black}they are} very easy to fail \cite{2019A, MaoFeng-401}. The unexpected failures of the bearing may cause a sudden shutdown of the machine, resulting in huge economic and time losses. Therefore, accurate and reliable bearing condition monitoring and fault diagnosis systems play an important role in preventing these failures. {\color{black}To date}, many bearing fault diagnosis methods have been reported in {\color{black}the relevant literature}. These fault diagnosis methods can be divided into two categories: methods based on signal processing techniques and the methods based on machine learning. Many fault diagnosis methods based on signal processing have been applied \cite{QinYi-396, ZhangWan-399, HuangWen-398, MangGao-400, LiWang-397, LeiLiu-395}. These methods can effectively filter {\color{black}noise} in signals and enhance the fault features. {\color{black}However, these methods require many manual operations with high professional knowledge.} Therefore, it is difficult to diagnose complex mechanical faults accurately and reliably in intelligent manufacturing, aerospace and other high-precision industries. In recent years, {\color{black}owing to} the improvement in information collection ability and computing power, intelligent data-driven fault diagnosis methods based on machine learning have received {\color{black}increasing attention} \cite{LiuYang-393, HongKim-394, 2018Fault, 2016Time}. {\color{black}In particular}, deep learning breaks the limitation that traditional machine learning needs to rely on specially designed features by virtue of its deep representation learning ability. It greatly improved the diagnostic accuracy and response speed of diagnosis methods with fewer requirements for professional knowledge. In recent years, deep-learning-based fault diagnosis {\color{black}methods have} been widely used and developed rapidly \cite{2018Machinery, 2019Simultaneous}. Huang et al. \cite{2019An} proposed an improved multiscale cascade convolutional neural network (MC-CNN). A new layer was added to the traditional convolutional neural network (CNN). The composed signal was obtained by concatenating the signals convolved by the original input and kernels of different lengths, which provided more useful information for the CNN and improved the model’s robustness to non-stationary conditions. Inspired by the dynamic routing capsule net, Zhu et al. \cite{Zhu2019A} proposed a capsule network with inception blocks and regression branches. This method achieved better generalization ability than state-of-the-art CNN models. Zhao et al. \cite{2020Deep} proposed a new deep learning method named ResNets with adaptive PReLUs (ResNet-APReLU), which assigns different nonlinear transformations to the input signal and improves the diagnostic accuracy. The recurrent neural network (RNN) and long short-term memory network (LSTM) have also been applied in the field of fault diagnosis. Shi et al. \cite{2019Rolling} proposed a fault diagnosis framework based on a stacked denoising auto-encoder (SDAE) and LSTM, which effectively detected the incipient fault of the rolling bearing. Qiao et al. \cite{Qiao2020Deep} illustrated a dual-input model based on the CNN and the LSTM. The features of the time and frequency domains were simultaneously used for end-to-end rolling bearing fault diagnosis under different noises and loads. In the cited literature, intelligent data-driven methods based on deep learning have been proven to be able to successfully solve conventional fault diagnosis tasks with {\color{black}sufficient} supervised training data \cite{2019Deep}.

Although the intelligent diagnosis methods proposed in the cited literature have shown good performance, the main limitation of the deep learning-based methods is {\color{black}increased needs of} a large number of high-quality labeled samples to train the model. {\color{black}Many} raw data {\color{black}must} be labeled in advance. However, in real industrial scenarios, {\color{black}because of} the uncertainty of systems and the high labor cost of labeling, condition monitoring data of machinery are usually recorded in a form of unlabeled samples. Owing to the lack of labeled samples, deep learning-based intelligent data-driven methods exhibit serious overfitting which affects the generalization performance of these methods and hinders their application in real industrial scenarios.

To solve the limitations of labeled data scarcity on the performance of deep learning-based intelligent diagnosis methods, researchers {\color{black}provided} many solutions. These methods can be roughly divided into transfer learning-based methods \cite{Hasan2019Acoustic, Tong2018Bearing}, data-augmentation-based methods \cite{Hu2019Data,Wang2020A} , semi-supervised learning-based methods \cite{Mao2020Online,Yu2020A} and unsupervised learning-based methods \cite{Liu2018A,Wei2017A}. Transfer learning attempts to transfer diagnostic knowledge from a source domain with sufficient labeled data to a target domain with insufficient labeled data. Considering the potential similarity between different rotating machines, Li et al. \cite{Li2020Diagnosing} proposed a transfer learning diagnosis method based on deep learning, which transferred the diagnosis knowledge learned from sufficient supervised data of multiple rotating machines to the target equipment with domain adversarial training. The experimental results based on the four datasets showed that it was feasible to improve the diagnostic performance by exploring different datasets. Based on basis of transfer learning, Guo et al. \cite{Guo2019Deep} proposed a deep convolutional transfer learning network (DCTLN) that includes a condition recognition module and a domain adaptation module. The effectiveness of the method was verified {\color{black}through} six fault-diagnosis experiments. Shao et al. \cite{Shao2019Highly} proposed a machine fault diagnosis framework based on a deep CNN, which used the transfer strategy to fine-tune a pre-trained deep network to improve the training efficiency of the model. This method achieved the most advanced results for the different datasets. The above results show that transfer learning can effectively relieve the problem of limited labeled samples in the target domain data. However, the successful use of transfer learning requires a large amount of high-quality labeled source domain data for training. {\color{black}However, it is costly to obtain these data.} Therefore, a method based on transfer learning cannot fundamentally solve this problem.

To solve the problem of insufficient labeled data, an intuitive method is to generate additional samples based on the available labeled data to augment the limited supervised dataset. Many data-augmentation-based methods have been applied in the field of fault diagnosis. For instance, Shao et al. \cite{Shao2019Generative} developed an auxiliary classifier generative adversarial network (ACGAN) based framework to learn diagnostic knowledge from mechanical sensor signals and generate realistic one-dimensional raw data. Li et al. \cite{Li2018Intelligent} proposed a bearing fault diagnosis method that included two data augmentation methods (sample-based method and data-based method) and five data enhancement techniques. The effectiveness of the method was verified using two rolling-bearing datasets. Meng et al. \cite{Meng2019Data} proposed a new data augmentation method based on original data, which decomposed a single sample into multiple monomers, and then recombined the monomers to increase the number of data samples. Although the training dataset can be significantly expanded by data augmentation, the augmented samples are artificially generated based on the basis of the existing labeled samples. Because the generated samples are similar to the original samples, they lack diversity and authenticity. Therefore, the improvement of the generalization ability is limited by using data-augmentation-based methods.

In actual industrial scenarios, although it is difficult to obtain high-quality labeled samples, there is a lot of unsupervised data available \cite{Liu2018Unsupervised}. These unsupervised data are collected under real working conditions of mechanical equipment, which contain rich information on machine operating conditions and health states. By making full use of unsupervised data, the generalization ability of intelligent data-driven models based on deep learning can be effectively improved \cite{Li2020Deep}. Semi-supervised learning can use both a limited number of labeled samples and a large number of unsupervised samples to train a discriminative model and improve the recognition accuracy. Therefore, it is a suitable method for fault diagnosis with insufficient labeled data. In recent years, {\color{black}increasing attention has} been paid to the research on semi-supervised learning in fault diagnosis. Li et al. \cite{Li2020Deep} proposed a three-stage training scheme, including pre-training, representation clustering and enhanced supervised learning. The proposed method was verified on two rotating machine datasets and achieved good performance on both semi-supervised and unsupervised learning tasks. Yu et al. \cite{Yu2021A} proposed a three-stage semi-supervised learning method based on data augmentation and metric learning. In the case of limited labeled samples, this method obtained better diagnosis results than the existing diagnostic methods on two datasets. Li et al. \cite{Li2019Semi} proposed a new augmented deep sparse auto-encoder (ADSAE) method. The research study showed that the ADSAE method effectively improved the generalization ability and robustness of the network model. Zhang et al. \cite{Zhang2019Fault} proposed a semi-supervised method of multiple association layer networks (SS-MALN) framework for fault diagnosis of planetary gearboxes. The SS-MALN model {\color{black}has the} advantages of semi-supervised learning and deep learning, and is superior to the traditional fault classification model based on deep networks in the case of less labeled data. The purpose of semi-supervised learning is to improve the generalization ability of the discriminant model with the help of unsupervised data. According to the semi-supervised assumptions (SSAs) proposed by Chapelle \cite{2006Semi}, semi-supervised learning can work only when the unlabeled data meet the requirements of smoothness estimation, cluster estimation and manifold estimation. Therefore, enhancing the clustering characteristics of unsupervised data, which makes the feature distribution of the same class of unsupervised data more closely, can improve the performance of semi-supervised learning. In the cited semi-supervised fault diagnosis methods, references \cite{Li2020Deep} and \cite{Yu2021A} used a combination of k-means and metric learning to enhance the clustering characteristics of unsupervised data. {\color{black}First}, the unsupervised data were clustered, and then the data of the same cluster were distributed more closely through metric learning. However, k-means and metric learning were two independent stages in this way, and the error caused by clustering would directly result in a negative impact on metric learning. Researching more accurate clustering algorithms is a way to overcome this shortcoming. In recent years, deep clustering which combines deep learning with clustering, has developed rapidly \cite{Aljalbout2018Clustering}. Deep clustering can significantly improve the clustering performance by learning the clustering-friendly representation of the original data through a deep neural network \cite{Min2018A}. Even though {\color{black}many} achievements have been made in {\color{black}several research studies} of deep clustering \cite{Mukherjee2018ClusterGAN,Madiraju2018Deep,guo2017improved,2015Unsupervised}, there are few applications in the field of bearing fault diagnosis. An et al. \cite{An2021Deep} introduced a deep clustering method based on autoencoded embedding and local manifold learning. Experiments on the Case Western Reserve University (CWRU) bearing dataset {\color{black}turned out} that this method could find the optimal clustering manifold and obtain better clustering results than the current advanced baseline methods. In this method, an auto-encoder is used for data dimensionality reduction, and manifold learning is used to learn the manifold structure of the data. {\color{black}Because of} the different learning purposes of each stage, the model could not achieve end-to-end training, {\color{black}and thus} the features learned by the auto-encoder might not be suitable for manifold learning. In addition, the reliability of unsupervised learning cannot be guaranteed without any labeled data or discriminative data \cite{Zhang2019Fault}. 

{\color{black}In this paper, a fault diagnosis method based on MS-SSIDEC is proposed to solve the problem of bearing fault diagnosis with limited labeled data. Unlike most semi-supervised learning methods and unsupervised learning methods, MS-SSIDEC combines the advantages of both of them. In real-world industries, there are scarce available labeled data and excessive unlabeled data. Although the use of clustering to generate pseudo labels is already a commonly leveraged method, the poor performance of traditional clustering methods may lead to a negative effect on fault diagnosis. A method that can effectively explore labeled and unlabeled data simultaneously is urgently needed. MS-SSIDEC introduces IDEC, the most advanced deep clustering approach, to mine unlabeled data. In addition, IDEC is further improved to fit the fault diagnosis task under the situation of insufficient labeled data. In the first stage, the SCCAE is proposed and pre-trained to automatically extract the fault features of unlabeled data. Because SCCAE employs the skip-connection \cite{2015Lateral} to connect the encoder and decoder, information loss caused by data compression as well as the pressure caused by low-dimensional features representing all information can be reduced. Therefore, the learned features contain more meaningful fault information. In the second stage, an SSIDEC model, which integrates the pre-trained auto-encoder with a clustering layer, is proposed to make the fault features learned by SCCAE more clustering-friendly and achieve clustering simultaneously. Compared to the traditional IDEC, SSIDEC introduces a small number of labeled samples as the constraint to ensure the correct initialization of clustering centers, which improves the performance of clustering and extends the IDEC to the semi-supervised learning. To reduce overfitting in the training of SSIDEC, VAT \cite{2018Virtual,2018RDEC} is introduced as a regularization term to improve the robustness of the model to local perturbations of data. In the third stage, the labeled dataset is augmented with the pseudo-labeled data obtained from the second stage and then utilized for training a bearing fault diagnosis model.}

{\color{black}By combining semi-supervised learning and deep clustering, MS-SSIDEC shows promising potential in both semi-supervised and unsupervised learning tasks for fault diagnosis. This is a applicational innovation for the field of bearing fault diagnosis in where semi-supervised and unsupervised tasks are usually solved independently. In semi-supervised tasks, available labeled data can be effectively used to initialize clustering centers of the method and augmented data are obtained by high-quality clustering of real industrial data, {\color{black}which are more realistic and diverse} than those generated by current artificial methods. Thus the generalization of the model trained by these augmented data can be furthur improved. Meanwhile, in purely unsupervised tasks, MS-SSIDEC benefited from by VAT and SCCAE can achieve better performance than traditional methods. The main contributions of our work can be summarized as follows:}

\begin{enumerate}[(1)]
\item In this paper, an MS-SSIDEC-based bearing fault diagnosis method is proposed to improve the performance of bearing fault diagnosis {\color{black}under the situation of insufficient labeled data} by combining semi-supervised learning and deep clustering. A large number of experiments on {\color{black}datasets provided by the CWRU and the Machinery Failure Prevention Technology (MFPT)} showed that the proposed method achieved good performance in both semi-supervised and unsupervised learning tasks and showed generalization ability under different working conditions.

\item A novel auto-encoder structure SCCAE is proposed to automatically learn the low-dimensional representations of the original frequency domain signals of bearings. Because skip-connection can effectively reduce the information loss caused by fusing multi-scale information between the encoder and decoder, the low-dimensional representations learned by SCCAE contain more meaningful fault information. 

\item A deep clustering method, SSIDEC, was proposed to enrich the limited labeled dataset. {\color{black}By imposing constraints on the initialization of cluster centroids with a small number of labeled samples and introducing VAT as a regularization term, SSIDEC achieves higher quality clustering results than traditional deep clustering. SSIDEC can mitigate the limitation on the generalization ability of the diagnostic model caused by insufficient labeled samples.}
\end{enumerate}

The rest of this paper is organized as follows. In the second section, the theoretical basis and process of the method are described in detail. In the third section, extensive experimental verification of {\color{black}two} bearing datasets is provided. Finally, the conclusions are drawn in the fourth part.

\section{Proposed method}
\label{sec:1}
\subsection{Problem fomulation}
\label{sec:2}
In this study, a semi-supervised bearing fault diagnosis problem with limited labeled data and sufficient unsupervised data is investigated. $D_{sp}=\{(x_i^{sp},y_i^{sp})\}_{i=1}^{n_{sp}}$ denotes the small set of labeled data, where $x_i^{sp}\in{\mathbb{R}^{N_{input}}}$ represents the sample data of $N_{input}$ dimensions, $y_i^{sp}$ is the health condition label of the equipment corresponding to the sample, and $n_{sp}$ is the number of samples. $D_{un}=\{x_i^{un}\}_{i=1}^{n_{un}}$ represents the unsupervised dataset, where $x_i^{un}\in{\mathbb{R}^{N_{input}}}$ represents the sample data and $n_{un}$ is the number of unsupervised samples.

The purpose of this {\color{black}study} is to build a {\color{black}fault-diagnosis} framework based on $D_{sp}$ and $D_{un}$. The framework has strong generalization ability on the test dataset $D_{test}=\{(x_i^{test},y_i^{test})\}_{i=1}^{n_{test}}$, where $x_i^{test}$ represents sample data, $y_i^{test}$ is the corresponding label, and $n_{test}$ is the number of test samples. To evaluate the robustness of the proposed method to changes in operating conditions, the data space $D_{sp}$, $D_{un}$, $D_{test}$ can be divided into different spatial distributions under the same label space.

The research {\color{black}presented} in this paper can also be used to solve unsupervised learning tasks in fault diagnosis. When there are only unsupervised data, the health conditions of the unsupervised data can be obtained directly by clustering.

\subsection{Framework}
\label{sec:3}

\begin{figure*}[htbp]
	\centering\includegraphics[width=0.55\textwidth]{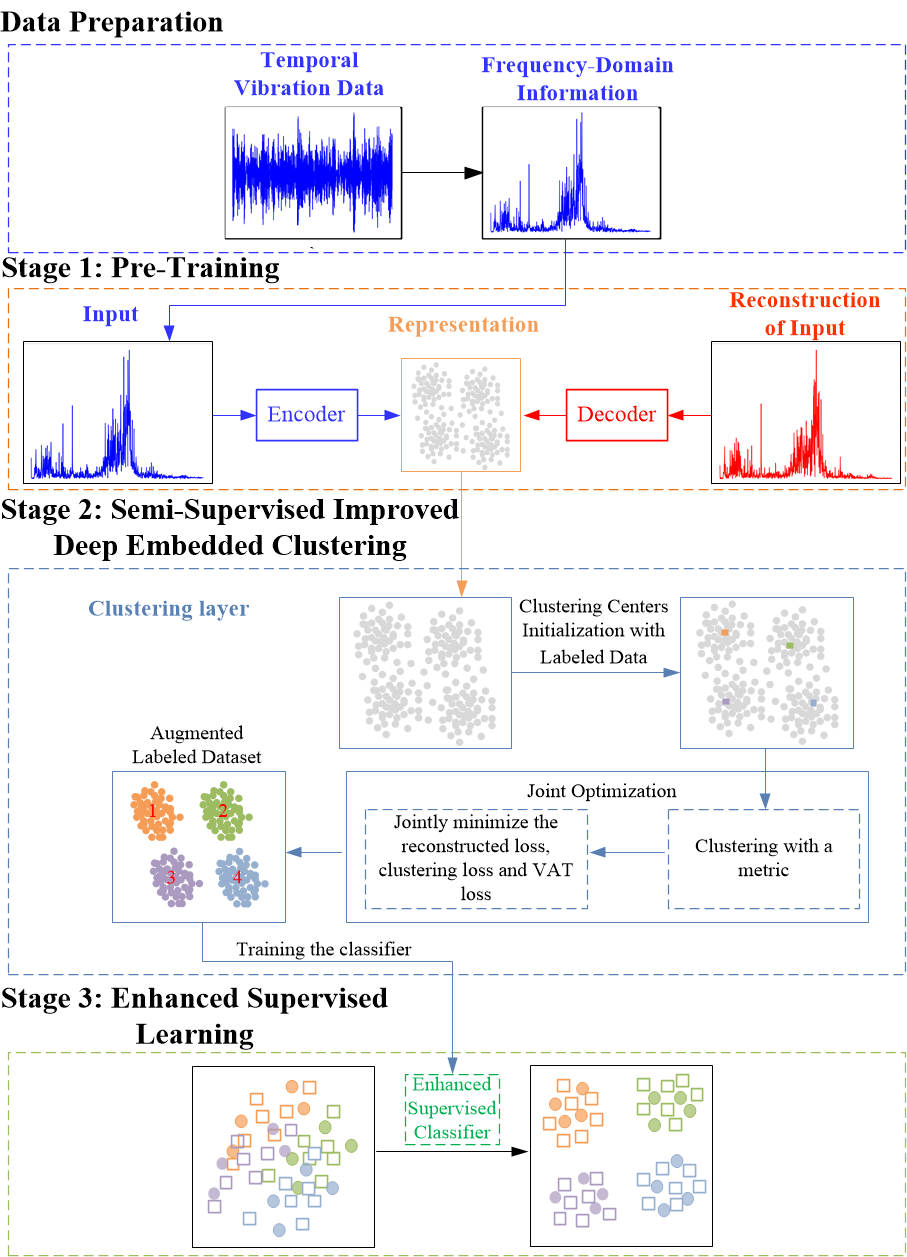}
	\caption{Framework of the proposed fault diagnosis method.}
	\label{fig:1}       
\end{figure*}

Fig. \ref{fig:1} {\color{black}shows the framework} of the proposed method. First, the original mechanical vibration acceleration signals are collected by sensors {\color{black}and then transformed into the frequency domain using a fast Fourier transform (FFT). These frequency-domain samples are used to be input data of the network.} The proposed fault diagnosis method consists of three stages: pre-training, semi-supervised improved deep embedded clustering, and enhanced supervised learning. {\color{black}In the pre-training stage, a novel auto-encoder is trained to learn low-dimensional representations of supervised and unsupervised data. In the second stage, an SSIDEC network that integrate a clustering layer and the auto-encoder is proposed to achieve the clustering assignment. Meanwhile, by minimizing the Kullback-Leibler (KL) divergence between soft assignments and auxiliary target distributions and preserving local structure of features, low-dimensional representations learned from the first stage can be more clustering-friendly. Finally, the labeled dataset can be enriched by pseudo-labeled samples obtained from the second stage and used to train a bearing fault diagnosis model.}

\subsection{Principle in stage 1: Pre-Training}
\label{sec:4}

\begin{figure*}
	\centering\includegraphics[width=0.75\textwidth]{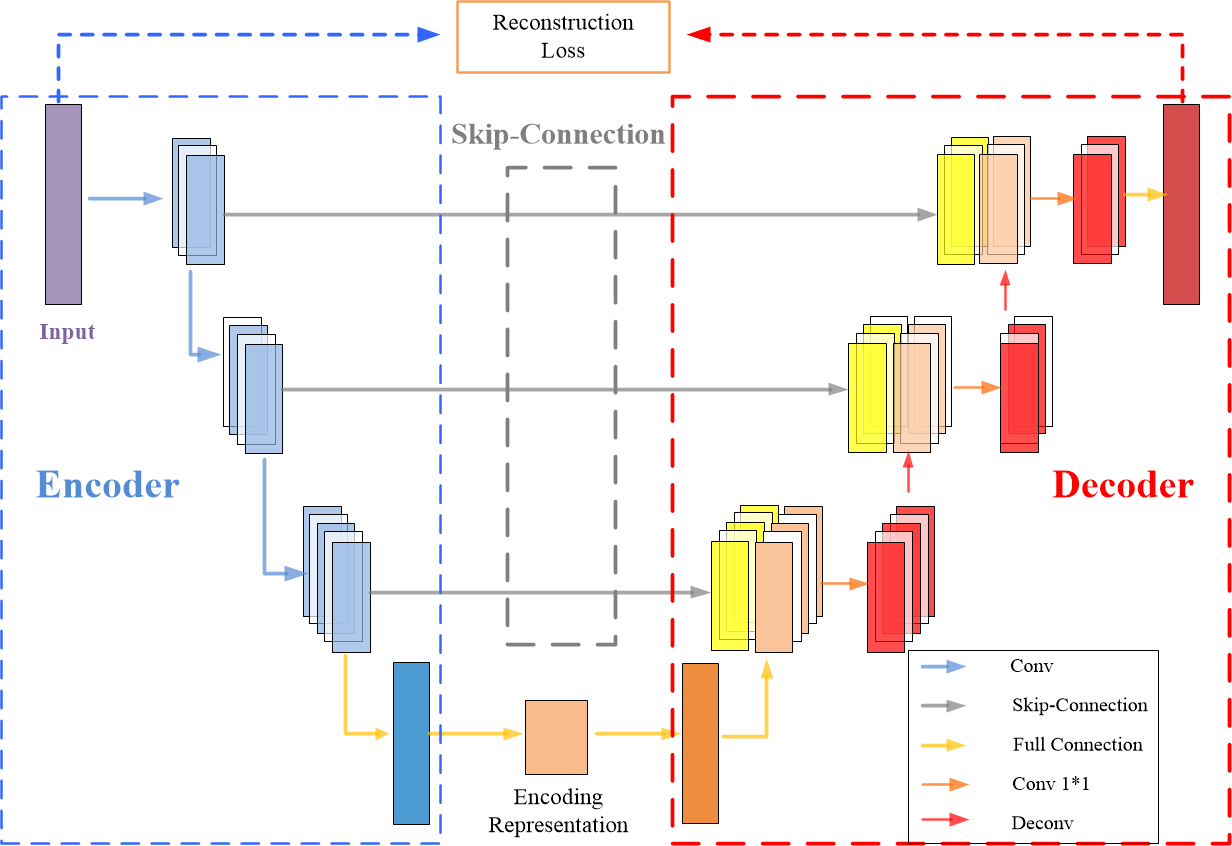}
	\caption{Architectures of the SCCAE in the first stage.}
	\label{fig:2}       
\end{figure*}

{\color{black}In stage 1, the proposed SCCAE is first pre-trained and used to extract low-dimensional fault features from labeled and unlabeled data.} As shown in Fig. \ref{fig:2}, the auto-encoder \cite{0Reducing} is a deep neural network architecture {\color{black}that includes an} encoder and a decoder, which can adaptively extract low-dimensional features of unsupervised input data and effectively avoid overfitting \cite{Li2019Semi}. {\color{black}The encoder maps the input data into a feature space in where main characteristics of these data can be represented by low-dimensional features. The decoder reconstructs the input data from thses features. There is a reconstruction error between the reconstructed and original data.} The smaller the reconstruction error, the more meaningful information can be maintained in low-dimensional representations. Specifically, the function of the encoder is to calculate the features $z_i=f_{en}(x_i)$ of $x_i$ in the space $Z$ {\color{black}using} a coding equation $f_{en}$ with a parameter of $\theta_{en}$. The decoder reconstructs the input data {\color{black}using} the decoding equation $f_{de}$ with a parameter $\theta_{de}$. In general, the auto-encoder model $f_{ae}$ can be expressed as,
\begin{align}
\hat{x}_i &=f_{ae}(x_i)=f_{de}(f_{en}(x_i)),\notag \\
z_i &=f_{en}(x_i),\notag\\
\hat{x_i} &=f_{de}(z_i), 
\end{align}
{\color{black}To further reduce the reconstruction error, the skip-connection is used to fuse high-level and low-level features between the corresponding layers of the encoder and the decoder. This connection directly flows the detailed information extracted by the encoder to the decoder, which supplements the details lost in the process of data compression, and can reduce the pressure of high-level low-dimensional features representing all information. The fusion features contain the multi-scale and multi-level information extracted from the shallow and deep layers of the auto-encoder, which reduce the reconstruction error and makes the auto-encoder model perform better in low-dimensional representation learning.} The auto-encoder optimizes the model parameters $\theta_{en}$ and $\theta_{de}$ by minimizing the reconstruction error $L_{rec}$ as,
\begin{align}
& L_{rec}=\frac{1}{n}\sum_{i=1}^{n_{pre}}{\Vert{x_i-\hat{x}}_i\Vert}^2, \notag \\
& \hat{\theta}_{en}, \hat{\theta}_{de}= \mathop{\arg\min}_{\theta_{en},\theta_{de}} L_{rec}(\theta_{en}, \theta_{de})
\end{align}
where $n_{pre}$ represents the number of samples used in the pre-training and usually contains a small {\color{black}amount} of supervised data and a large {\color{black}amount} of unsupervised data. $\hat{\theta}_{en}$ and $\hat{\theta}_{de}$ represent the optimal values of $\theta_{en}$ and $\theta_{de}$, respectively. The optimization problem can be solved {\color{black}using the} stochastic gradient descent algorithm, and the parameters can be optimized as,
\begin{align}
&\theta_{en}\gets\theta_{en}-\delta\frac{\partial{L_{rec}}}{\partial{\theta_{en}}}, \notag \\
&\theta_{de}\gets\theta_{de}-\delta\frac{\partial{L_{rec}}}{\partial{\theta_{de}}},
\end{align}
where $\delta$ represents the learning rate.

\subsection{Principle in stage 2: Semi-Supervised Improved Deep Embedded Clustering}
\label{sec:5}
{\color{black}In stage 2, the SSIDEC method is proposed to achieve semi-supervised learning and unsupervised learning. IDEC can simultaneously perform clustering and learn clustering-friendly features by constraining the reconstruction and clustering losses.} The network model includes a pre-trained auto-encoder and a clustering layer containing $N_{cluster}$ clustering centers $\left\{\mu_{j}\in{Z}\right\}_{j=1}^{N_{cluster}}$, and the learning process can be divided into two stages: initialization of clustering centers and joint optimization.

In the IDEC network model, the initialization of the cluster center is important because clustering centers directly reflect the potential representation of data \cite{Madiraju2018Deep}. To ensure that the initial center of each cluster is correct and extend IDEC to semi-supervised learning, the SSIDEC method, which consists of two different initialization modes, is proposed. In the semi-supervised learning task, the centroids of the low-dimensional features $f_{en}\left(x_i^{sp}\right)$ of $x_i^sp$ can be used as the initialization clustering centers directly. The initial clustering centers can be calculated as,
\begin{align}
&\mu_k=\hat{z}_i^{sp,k}, \notag \\
&z_i^{sp,k}=f_{en}(x_i^{sp}|y_i^{sp}=k), \notag \\
&k=1,2,...,N_{cluster}, \label{eq4}
\end{align}
where $z_i^{sp,k}$ is the low-dimensional representation set of labeled samples belonging to $k$ class, and $\hat{z}_i^{sp,k}$ is the average of $z_i^{sp,k}$. Eq. \ref{eq4} shows that in the semi-supervised problem, clustering centers are initialized using the mean values of the low-dimensional representations of the labeled data in each class.

In the unsupervised problem, only unlabeled data $D_{un}=\left\{x_i^{un}\right\}_{i=1}^{n_{un}}$ are available. Initial centers $\left\{\mu_j\in{Z}\right\}_{j=1}^{N_{cluster}}$ are determined by clustering results of features $f_{en}\left(x_i^{un}\right)$ of $x_i^{un}$ using k-means.

After determining the initial clustering center $\mu_j$, a two-step alternating unsupervised algorithm is used for joint optimization. First, the probability $q_{ij}\left(q_{ij}\in{Q}\right)$ is calculated where the feature $z_i$ belongs to the $j$ cluster center. The closer {\color{black}it is} to the cluster center $\mu_j$, the higher is the probability $q_{ij}$. The probability between each low-dimensional representation and each clustering center is measured by the student's t-distribution. The calculation method can be defined as,
\begin{align}
q_{ij}=\frac{\left(1+\frac{siml\left(z_i,\mu_j\right)}{\alpha}\right)^{-\frac{\alpha+1}{2}}}{\sum_{j}\left(1+\frac{siml\left(z_i,\mu_j\right)}{\alpha}\right)^{-\frac{\alpha+1}{2}}},
\end{align}
where $\alpha$ is the degree of freedom of student's t-distribution. $siml\left(z_i,\mu_j\right)$ is a similarity measurement method that is used to calculate the distance between the embedded sample $z_i$ and the cluster center $\mu_j$. To simplify the model, the Euclidean distance is used. $q_{ij}$ is a soft assignment, and $Q$ is the prediction distribution.

Then, an auxiliary target distribution $P$ is defined for $Q$. The calculation method of $p_{ij}\left(p_{ij}\in{P}\right)$ can be defined as,
\begin{align}
p_{ij}=\frac{q_{ij}^2/\sum_{i}q_{ij}}{\sum_{j}\left(q_{ij}^2/\sum_{i}q_{ij}\right)}.
\end{align}

The purpose of $q_{ij}^2$ is to attract data closer to the clustering center, so the auxiliary target distribution can strengthen the high-confidence prediction. {\color{black}Simultaneously}, the loss contribution of each center can be normalized to prevent large clusters from distorting the hidden feature space.

After obtaining the auxiliary target distribution, the clustering loss $L_c$ that is a KL divergence loss between the soft assignment $Q$ and the auxiliary target distribution $P$ can be defined as,
\begin{align}
L_c=KL\left(P\left|\right|Q\right)=\sum_{i}\sum_{j}p_{ij}log\frac{p_{ij}}{q_{ij}}.
\end{align}
By minimizing $L_c$, the soft assignment can be gradually matched to the auxiliary target distribution. Therefore, SSIDEC can learn more clustering-friendly features of unsupervised data.

Overfitting is a common problem in unsupervised learning. {\color{black}VAT is introduced into the training stage as a regularization term to reduce overfitting and improve the generalization ability of the model. VAT adds perturbations in a specific direction to the input data for data augmentation, which improves the invariance of the model to the label prediction distribution. This regularization method ensures that the data of the same class in the original space maintain a similar distribution in the feature space, and thus benefits the clustering task.} $L_v$, the loss of VAT, which is the KL divergence loss between the predicted distribution of the original data $x$ and the augmented data $x+r_{adv}$, can be defined as,
\begin{align}
L_v=KL(Q\left|\right|Q\left(x+r_{adv}\right)),
\end{align}
{\color{black}where $Q$ is the predicted distribution, and $r_{adv}$ is the adversarial perturbation which can be calculated in an adversarial manner as,}
\begin{align}
r_{adv}=\mathop{\arg\max}_{r;\Vert{r}\Vert\leq\varepsilon}KL(Q\left|\right|Q\left(x+r\right)),
\end{align}
where $r$ is a perturbation that does not alter the meaning of the data point, and $\varepsilon$, a hyperparameter, is the perturbation size. {\color{black}Perturbation size $\varepsilon$ specifies the range of nearby points requiring consideration during learning. Its effects have been investigated in reference \cite{2018Virtual} in where VAT achieves good performance when $\varepsilon$ is fixed as 2. Therefore, according to the suggestion, $\varepsilon=2$ in this study.}

Preserving the local structure of the data can improve the performance of the clustering methods \cite{guo2017improved}. The auto-encoder can learn low-dimensional representations with the unique local structure of the input data by minimizing the reconstruction loss $L_{rec}$, which constrains the distortion of the embedding space caused by $L_c$. Therefore, the decoder network in the auto-encoder structure is retained in the proposed clustering network model.

In summary, SSIDEC combines reconstruction loss $L_{rec}$, clustering loss $L_c$, and VAT loss $L_v$ for joint optimization. The objective function can be defined as,
\begin{align}
&L=L_{rec}+\gamma_cL_c+\gamma_{vat}L_v, \notag \\
&\hat{\theta}_{en},\hat{\theta}_{de},\hat{\mu}_j=\mathop{\arg\max}_{\theta_{en},\theta_{de},\mu_j}L\left(\theta_{en},\theta_{de},\mu_j\right),
\end{align}
where $\gamma_c>0$ is the clustering loss coefficient, which is used to control the degree of distortion of the low-dimensional feature space. {\color{black}$\gamma_{vat}$ is the regularization coefficient used to control the effect of VAT. It can be fixed as 1 according to the experimental results of reference \cite{2018Virtual}.} The update of the network model parameters can be defined as,
\begin{align}
&\mu_j\gets\mu_j-\delta\left(\gamma_c\frac{\partial{L_c}}{\partial{\mu_j}}+\gamma_{vat}\frac{\partial{L_v}}{\partial{\mu_j}}\right), \notag \\
&\theta_{de}\gets\theta_{de}-\delta\frac{\partial{L_{rec}}}{\partial{\theta_{de}}}, \notag \\
&\theta_{en}\gets\theta_{en}-\delta\left(\frac{\partial{L_{rec}}}{\partial{\theta_{en}}}+\gamma_c\frac{\partial{L_c}}{\partial{\theta_{en}}}+\gamma_{vat}\frac{\partial{L_v}}{\partial{\theta_{en}}}\right),
\end{align}
where $\delta$ represents the learning rate.

The low-dimensional representation $z_i$, clustering center $\mu_j$, and soft assignment $q_{ij}$ of the input data are updated in each epoch. In addition, the auxiliary target distribution is updated every $tau$ epochs. After each iteration, a pseudo-label $s_i$ is assigned to each sample $x_i$,
\begin{align}
s_i=\mathop{\arg\max}_{j}q_{ij}.
\end{align}

When the change rate of the pseudo-label of all samples between two consecutive updates of the auxiliary target distribution is less than the threshold $tol\%$ or the training reaches the maximum number of iterations $Itr_{max}$, the learning process is stopped. The detailed algorithm of the second stage is summarized in Algorithm 1.

\begin{algorithm}[!h]
	\caption{SSIDEC with VAT regularization} 
	\hspace*{0.02in} {\bf Input:} 
	Input data: $X$; Number of clusters: $K$; Target distribution update interval: $\tau$; \\
	\hspace*{0.02in} Stopping threshold: $tol\%$; Maximum iterations: $Itr_{max}$.\\
	\hspace*{0.02in} {\bf Output:} 
	Auto-encoder’s weights $\theta_{en}$ and $\theta_{de}$; Cluster centers $\mu$ and pseudo-labels $s$.
	
	\begin{algorithmic}[1]
		\State Initialize $\mu$, $\theta_{en}$ and $\theta_{de}$ according to (2) and (4) 
		\For{$iter\in{\left\{0,1,...,Iter_{max}\right\}}$} 
		\If{$iter\%T==0$} 
		\State Compute all embedded points $\left\{z_i=f_{en}\left(x_i\right)\right\}_{i=1}^n$. 
		\State Updata $P$ using (5), (6) and $\left\{z_i\right\}_{i=1}^n$.
		\State Save last pseudo-label assignment: $s_{old}=s$.
		\State Compute new pseudo-label assignments $s$ via (12).
		\If{$sum\left(s_{old}\neq{s}\right)/n<tol{\%}$}
		\State Stop training.
		\EndIf
		\EndIf
		\State Choose a batch of samples $S\in{X}$.
		\State Update $\mu$, $\theta_{en}$ and $\theta_{de}$ via (11) and $S$.
		\EndFor
	\end{algorithmic}
\end{algorithm}

\subsection{Principle in stage 3: Enhanced Supervised Learning}
\label{sec:6}

\begin{figure*}
	\centering\includegraphics[width=0.75\textwidth]{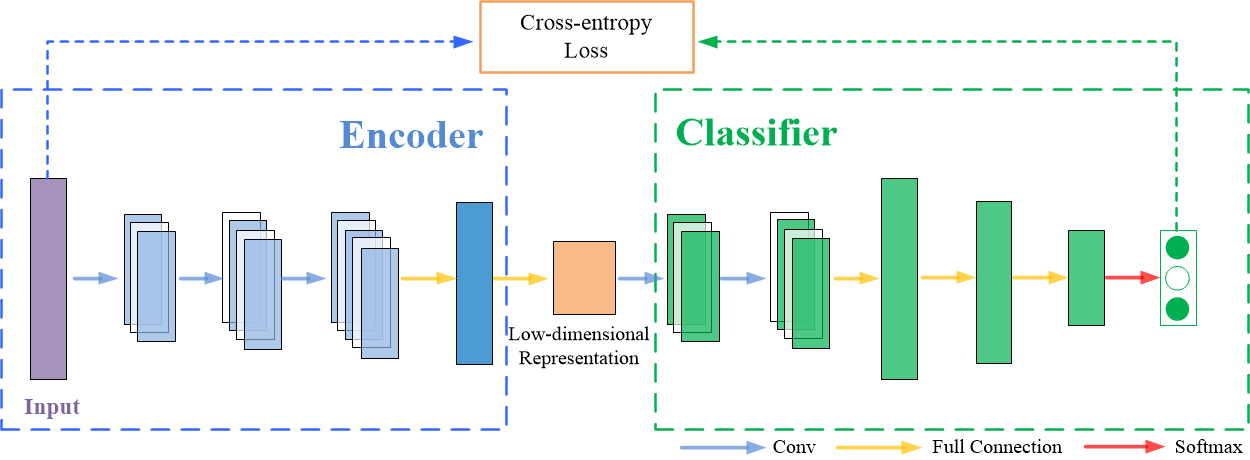}
	\caption{Architectures of the discrimination model in the third stage.}
	\label{fig3}       
\end{figure*}

{\color{black}After stage 2, unlabeled data can be well clustered into the corresponding groups of the labeled data, and the labeled dataset can be then effectively augmented by labeling pseudo labels. A bearing fault discriminative model is shown in Fig. \ref{fig3}. This model consists of an encoder and a classifier and can be well trained using the augmented dataset.} The encoder shares same weights with the encoder of the proposed SCCAE in stage 1. The clustering-friendly representation of input data extracted by it is then recognized by the classifier. In this stage, the cross-entropy loss for the objective function can be defined as,
\begin{align}
L_s=-\frac{1}{n_{sp}}\sum_{i=1}^{n_{sp}}\sum_{k=1}^{N_{cluster}}1\left\{y_i^{sp}=k\right\}\log\frac{e^{\chi_{c,i,k}^{sp}}}{\sum_{j=1}^{N_{cluster}}e^{\chi_{c,i,k}^{sp}}} \notag \\-\frac{1}{n_{un}}\sum_{i=1}^{n_{un}}\sum_{k=1}^{N_{cluster}}1\left\{y_i^{un}=k\right\}\log\frac{e^{\chi_{c,i,k}^{un}}}{\sum_{j=1}^{N_{cluster}}e^{\chi_{c,i,k}^{un}}},
\end{align}
where $\chi_{c,i,k}^{sp}$ and $\chi_{c,i,k}^{un}$ represent the $k$-th vector in the output obtained by taking the $i$-th supervised sample and the unsupervised sample as the input of the discrimination model. By minimizing $L_s$, the parameters of the discriminative model can be updated as,
\begin{align}
&\hat{\theta}_{en},\hat{\theta}_c=\mathop{\arg\min}_{\theta_{en},\theta_c}L_s\left(\theta_{en},\theta_c\right), \notag \\
&\theta_c\gets\theta_c-\delta\frac{\partial{L_s}}{\partial{\theta_c}}, \notag \\
&\theta_{en}\gets\theta_{en}-\delta\frac{\partial{L_s}}{\partial{\theta_{en}}},
\end{align}
where $\theta_{en}$ is the parameters of encoder and $\theta_c$ is the parameters of clasiffier; $\hat{\theta}_{en}$ and $\hat{\theta}_c$ are the optimal parameters of $\theta_{en}$ and $\theta_c$, respectively, and $\delta$ is the learning rate.

\subsection{The general procedure of the proposed method}
\label{sec:7}

\begin{figure*}[htbp]
	\centering\includegraphics[width=0.95\textwidth]{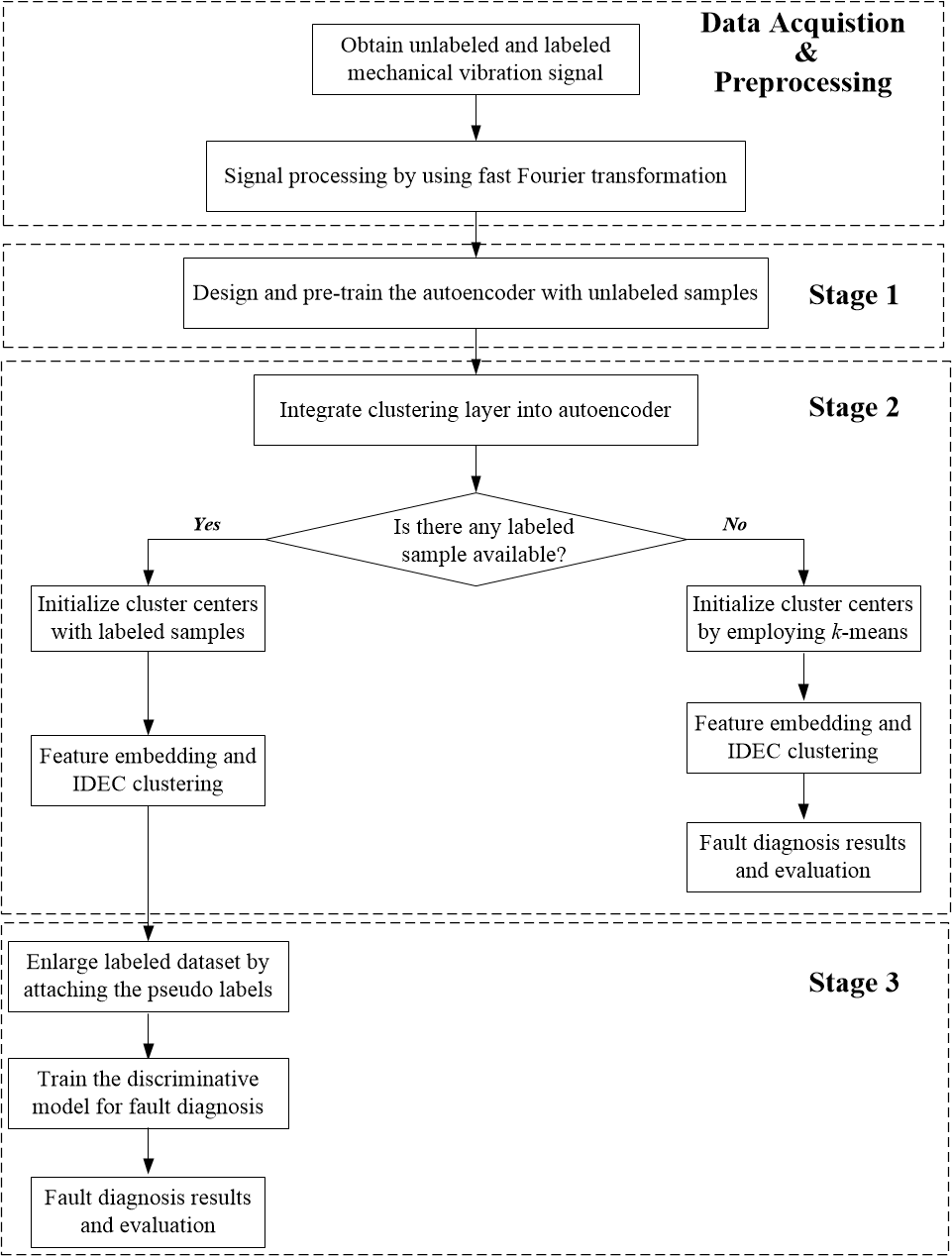}
	\caption{Information flow of the proposed fault diagnosis method.}
	\label{fig4}       
\end{figure*}
{\color{black}To address the problem of bearing fault diagnosis in real industrial scenarios in where limited labeled and sufficient unlabeled data are available,} an MS-SSIDEC method is proposed. This method expands the labeled dataset through SSIDEC and thus improves the generalization performance of semi-supervised learning. By improving the initialization of clustering centers, these labeled samples are used to facilitate the accuracy of deep clustering. VAT regularization is used in the training process to avoid overfitting. In addition, this method can directly solve the problem of unsupervised fault diagnosis. The detailed process of this method is illustrated in Fig. \ref{fig4}. First, the data are pre-processed, and the frequency-domain information of the unsupervised vibration signal of the bearing is obtained through FFT. Then, a three-stage strategy is adopted. In the first stage, the proposed SCCAE network model is pre-trained. In the second stage, an SSIDEC model is proposed to achieve the clustering assignment. According to whether the diagnostic task is unsupervised or semi-supervised, the cluster centers are initialized in separately modes, and thus the proposed method is extended to both semi-supervised learning and unsupervised learning tasks. In the third stage, for the semi-supervised bearing fault diagnosis task, the pseudo-labeled samples obtained by clustering are used to expand the labeled dataset to train the bearing fault discriminative model.

\section{Experimental study}
\label{sec:8}
\subsection{Data description}
\subsubsection{CWRU dataset}
\label{sec:9}

\begin{figure}[htbp]
	\vspace{-0.8cm}
	\centering\includegraphics[width=0.5\textwidth]{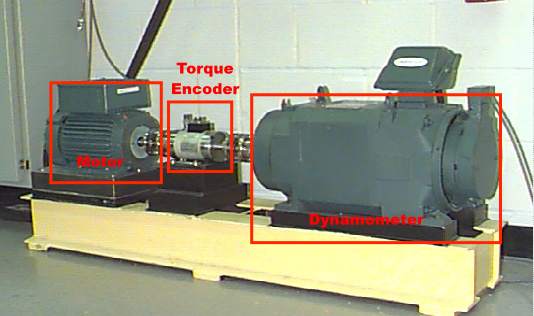}
	\caption{Test-rig of the rolling bearing.}
	\label{fig:5}       
\end{figure}

In this study, the data provided by the CWRU bearing data center \cite{Wade2015Rolling} is used to verify the proposed method. The test rig is shown in Fig. \ref{fig:5}, including a 2 hp motor, a torque encoder, a dynamometer, and control electronics. The bearing data are collected {\color{black}using} an acceleration sensor with a sampling frequency of 12kHz. {\color{black}The dataset contains bearing failure data under four working conditions with loads of 0, 1, 2 and 3 hp, and speeds of 1797, 1772, 1750 and 1730 rpm are implemented correspondingly.} 

\begin{table}[htpb]
	\footnotesize
	\centering
	\vspace{-0.4cm}
	\newcommand{\tabincell}[2]{\begin{tabular}{@{}#1@{}}#2\end{tabular}}
	\caption{Classes for the CWRU dataset.}
	\label{tab:1}       
	\begin{tabular}{lllllll}
		\hline\noalign{\smallskip}
		Label & Failure level & \tabincell{l}{Fault Size \\ (mil)} & \tabincell{l}{Load A\\(hp)} & \tabincell{l}{Load B\\(hp)} & \tabincell{l}{Load C\\(hp)} & \tabincell{l}{Load D\\(hp)}  \\
		\noalign{\smallskip}\hline\noalign{\smallskip}
		0 & Minor Ball Fault & 7 & 0 & 1 & 2 & 3 \\
		1 & Medium Ball Fault & 14 & 0 & 1 & 2 & 3 \\
		2 & Serious Ball Fault & 21 & 0 & 1 & 2 & 3 \\
		3 & Minor Inner Race Fault & 7 & 0 & 1 & 2 & 3 \\
		4 & Medium Inner Race Fault & 14 & 0 & 1 & 2 & 3 \\
		5 & Serious Inner Race Fault & 21 & 0 & 1 & 2 & 3 \\
		6 & Minor Outer Race Fault & 7 & 0 & 1 & 2 & 3 \\
		7 & Medium Outer Race Fault & 14 & 0 & 1 & 2 & 3 \\
		8 & Serious Outer Race Fault & 21 & 0 & 1 & 2 & 3 \\
		9 & Normal & 0 & 0 & 1 & 2 & 3 \\
		\noalign{\smallskip}\hline
	\end{tabular}
\end{table}

The vibration data used in this study were collected on bearings at the drive end of the motor and include four types of failures: (1) normal operating conditions (H); (2) outer ring failure (OF); (3) inner ring failure (IF) ; (4) ball failure (BF). Each type of fault has three different severity levels, represented by different fault diameters of 7, 14 and 21 mils, respectively. {\color{black}In summary}, the dataset contains a total of ten bearing health states under four operating conditions. The specific information is shown in Table \ref{tab:1}.

\subsubsection{MFPT dataset}

\begin{table}[htpb]
	\footnotesize
	\vspace{-0.4cm}
	\centering
	\newcommand{\tabincell}[2]{\begin{tabular}{@{}#1@{}}#2\end{tabular}}
	\caption{Classes for the MFPT dataset.}
	\label{tab:mfpt}       
	\begin{tabular}{ll}
		\hline\noalign{\smallskip}
		Label & Location of the fault  \\
		\noalign{\smallskip}\hline\noalign{\smallskip}
		0 & Normal baseline \\
		1 & Inner race ring \\
		2 & Outer race ring \\
		\noalign{\smallskip}\hline
	\end{tabular}
\end{table}

{\color{black}This dataset was provided by the MFPT society \cite{bechhoefer2016quick}. A test rig equipped with a NICE bearing was used to gather accerlerometer data for three conditions, normal baseline conditions (N) at 270 lbs of load and a sampling rate of 97,656 Hz, ten outer-raceway (OR) and seven inner-raceway (IR) fault conditions. Specifically, three outer-raceway faults were gathered with a load of 270 lbs and a sampling rate of 97,656 Hz and seven outer-raceway faults were assessed at varying loads: 25, 50, 100, 150, 200, 250 and 300 lbs. The sampling rate for these seven faults was 48,828 Hz. Seven inner race faults were analyzed at varying loads of 0, 50, 100, 150, 200, 250, and 300 lbs. The sampling rate for the inner race faults was 48,828 Hz. The specific information is shown in Table \ref{tab:mfpt}.}

\subsection{Implementation details of the network}
\label{sec:10}
In this study, the dimension of samples of bearing vibration acceleration data is set to 1024, and thus the dimension of the input data of the network obtained by the FFT is 512. Considering that the input data are one-dimensional signals in the frequency domain, one-dimentional CNN (1D-CNN) is selected as the basic network. In the first stage, the auto-encoder is pretrained. The encoder module contains three convolutional layers, the number of channels is 8, 16, and 32, the kernel size is set to 10, and the stride is set to 2. {\color{black}This architecture incorporates the guidelines proposed in references \cite{Li2020Deep,An2021Deep}, and some adjustments based on exhaustive model exploration were made to provide superior results for semi-supervised and unsupervised tasks of bearing fault diagnosis in this study.} After the convolutional layer, a flatten layer is set to fuse the local features into global features. After the flatten layer, the data {\color{black}are} embedded into the low-dimensional feature space through a fully-connected layer with the number of $N_{rep}$ neurons. {\color{black}The number of neurons in this layer is set to 32 in both weakly supervised and unsupervised tasks, and its effect on the model is investigated in Section \ref{sec:18}. The decoder structure is similar with that of the encoder including three fully-connected layers that is inverse to that of the encoder and three deconvolution modules with channels of 32, 16, 8, the kernel size of 10, and the stride of 2.} {\color{black}Owing} to the skip-connection, each deconvolution module of the decoder contains two parts. The first part concatenates the high-level and low-level features in the channel dimension to obtain a new feature. The second part performs 1*1 convolution on concatenated features to ensure that the dimensions of the corresponding layer features of the decoder and encoder are consistent. Finally, a reconstructed sample is obtained through a fully-connected layer with the number of $N_{input}$ neurons.

In the third stage, the discriminative model is trained. The discriminative model consists of two parts: an encoder and a classifier. {\color{black}Fellowing \cite{Li2020Deep}, the encoder shares the weight with the encoder in the pre-trained auto-encoder network model. The classifier contains two convolutional layers, a flatten layer, and two fully-connected layers. The number of channels of the convolutional layer is 8 and 16, the kernel size is set to 3, and the stride is set to 1. The two fully-connected layers contain 128 and $N_{cluster}$ neurons respectively. Those parameters were selected to handel the dataset used in the paper through exhaustive model exploration.} Finally, the softmax function {\color{black}is used} to classify the bearing health status.

\begin{table}[htpb]
	\centering
	\caption{Parameters used in this paper.}
	\label{tab:2}       
	\begin{tabular}{llll}
		\hline\noalign{\smallskip}
		Parameter & Value & Parameter & Value  \\
		\noalign{\smallskip}\hline\noalign{\smallskip}
		Epochs at Stage 1 & 4000 & $\gamma_{vat}$ & 1 \\
		Epochs at Stage 2 & 100 & $T$ & 20 \\
		Epochs at Stage 3 & 4000 & $tol\%$ & 1e-4 \\
		Learning rate at Stage 1 & 0.001 & $N_{input}$ & 512 \\
		Learning rate at Stage 2 & 1e-4 & $N_{rep}$ & 32 \\
		Learning rate at Stage 3 & 1e-4 & $N_{cluster}$ & 10 \\
		$\alpha$ & 1 & Batch size & 32 \\
		$\varepsilon$ & 2 &   &   \\
		\noalign{\smallskip}\hline
	\end{tabular}
\end{table}

{\color{black}In the model training process}, the Xavier normal initializer is used for parameter initialization. The convolutional layers are regularized by batch normalization (BN) \cite{ioffe2015batch}, and Leaky Relu \cite{2013Improving} is used as the activation function. All parameters are updated using the back-propagation algorithm, and adaptive moment estimation (Adam) \cite{2014Adam} is used as the optimizer. {\color{black}These selections are based on experience and experimental validations.}The maximum value of the pre-training epochs is set to 4000. In each training epoch, a mini-batch is used for gradient optimization, and the number of batches is set to 32 by default. {\color{black}The architecture of the network and parameters are determined based on validations in task C1 with $n_{sp}=1$ and $n_{un}=300$. }The experimental results in this paper adopt the average value of 10 trials, and the default values of the main parameters are listed in Table \ref{tab:2}.

\subsection{Performance with weakly supervised data}
\label{sec:11}
\subsubsection{Experimental setup}
\label{sec:12}

\begin{table}[htbp]
	\centering
	\caption{Data descriptions of the weakly supervised fault diagnosis tasks in the CWRU dataset (load).}
	\label{tab:3}       
	\begin{tabular}{llll}
		\hline\noalign{\smallskip}
		Task & Supervised data & Unsupervised data & Testing data  \\
		\noalign{\smallskip}\hline\noalign{\smallskip}
		C1 & 0 hp & 0 hp & 0 hp \\
		C2 & 0 hp & 0 hp & 1 hp,2 hp,3 hp \\
		C3 & 0 hp & 1 hp,2 hp,3 hp & 1 hp,2 hp,3 hp \\
		C4 & 0 hp & 0 hp,1 hp,2 hp,3 hp & 1 hp,2 hp,3 hp \\
		C5 & 2 hp & 2 hp & 2 hp \\
		\noalign{\smallskip}\hline
	\end{tabular}
\end{table}

{\color{black}First, the problem of bearing fault diagnosis using weakly supervised data is studied in the CWRU and MFPT datasets. Five diagnostic tasks in the CWRU dataset are listed in Table \ref{tab:3}. The numbers (0, 1, 2, 3) in the table represent four different working conditions (0, 1, 2, 3 hp).} To verify the performance of the proposed method under extreme conditions, {\color{black}in task C1, C2, C3 and C4,} there is only 1 labeled sample for each health condition i.e. $n_{sp}=1$, 300 unlabeled samples for each failure type i.e. $n_{un}=300$, and 300 testing samples for each failure type i.e. $n_{test}=300$. To verify the generalization ability of this method, the unsupervised samples contains data generated from varying bearing operating conditions. {\color{black}The task C5 is set to conduct the comparison of the proposed method and two generative-model-based methods. Because these generative-model-based methods require relatively large amount of labeled data and training samples, the number of total training samples is 10,808 and 1 \% of them i.e. $n_{sp}=108$ are labeled in this task.} {\color{black}For the MFPT dataset, one task, M1, is conducted in this study. In task M1, the total samples is as follows: N with 3,423, IR with 1,981, and OR with 5,404. 50\% of the total samples are used as training set and 1\% of them are labeled. Both training and testing sets consist of data under varying conditions.}

In this study, different methods are used for comparison:
\begin{enumerate}[(1)]
	\item CNN 
		
	In order to verify that the method proposed in this paper has better accuracy and generalization ability in the case of weak samples than the traditional method, a traditional CNN method is set for comparison. The traditional CNN method follows the traditional supervised learning paradigm, and the network structure is the same as {\color{black}that of} discriminative model of the proposed method. In this method, data augmentation or clustering methods are not used for sample generation. The model is pre-trainied by a large number of unsupervised samples and then fine-tuned by a small number of labeled samples.
	
	\item NoIDEC
	
	To reflect {\color{black}the fact} that the stage 2 can improve the clustering characteristics of unsupervised data and help to improve the performance of the semi-supervised learning fault diagnosis method, the NoIDEC method is set up for comparison and verification. In this method, without going through the SSIDEC training process, the initial clustering results are directly assigned as pseudo-label samples for data augmentation, and then enhanced supervised learning is performed.
	
	\item NoVAT
	
	To verify that the VAT regularization method can effectively reduce model overfitting and improve the generalization ability of the model. The NoVAT method is used for comparison. The Stage1 and Stage3 processes are consistent with the proposed method, and VAT regularization is not introduced in the training process of the second stage.
	
	\item DA \cite{2019Unsupervised}
	
	Data augmentation methods{\color{black}, which are typical semi-supervised learning methods,} are often used to solve semi-supervised learning problems in the field of fault diagnosis. In this study, a state-of-the-art data augmentation method, unsupervised data augmentation (UDA) \cite{2019Unsupervised}, is used for comparison.
	
	\item VAE \cite{san2019deep}
	
	{\color{black}The deep variational auto-encoder (VAE) is an effective approach for dimensionality reduction. Recently, a VAE-based bearing fault diagnosis method is reported in reference \cite{san2019deep}. For comparison with this advanced method, an experiment is conducted in task C5 and M1. The network structure and hypermeters are set according to the suggestion of reference \cite{san2019deep} and input data are preprocessed into spectrograms.} 
	
	\item GAN \cite{verstraete2020deep}
	
	{\color{black}In the reference \cite{verstraete2020deep}, a GAN based fault diagnosis method is reported proposed. This is a kind of semi-supervised learning method. To compare with this advanced method, an experiment is conducted in task C5 and M1. The network and parameters are set according to the reference \cite{verstraete2020deep} and input data are preprocessed into spectrograms.}

\end{enumerate}

\subsubsection{Diagnosis result}
\label{sec:13}

\begin{table}[htbp]
	\centering
	\caption{Average testing accuracies of the fault diagnosis tasks with weakly supervised data in the CWRU dataset.}
	\label{tab:4}       
	\begin{tabular}{llllll}
		\hline\noalign{\smallskip}
		Task & CNN(\%) & NoIDEC(\%) & NoVAT(\%) & DA(\%) & Proposed(\%) \\
		\noalign{\smallskip}\hline\noalign{\smallskip}
		C1 & 88.5 & 97.9 & 98.5 & 99.9 & 99.9 \\
		C2 & 64.5 & 73.2 & 89.8 & 87.5 & 90.5 \\
		C3 & 64.5 & 89.6 & 94.5 & 87.5 & 95.9 \\
		C4 & 64.5 & 84.1 & 92.8 & 87.5 & 93.7 \\
		\noalign{\smallskip}\hline
	\end{tabular}
\end{table}

In this section, the performance of the proposed MS-SSIDEC method for bearing fault diagnosis under weakly supervised conditions is verified. The diagnostic results in the CWRU dataset are presented in Table \ref{tab:4}. It can be seen that the proposed method can obtain better results than the commonly used semi-supervised learning methods and traditional supervised learning methods. It can be found that these methods, other than the traditional CNN method using supervised learning, have obtained higher test accuracy in task C1 as the labeled data and test data in this task are collected under the same bearing operating conditions. However, when the training and testing data are collected under different operating conditions, the performance of different diagnostic methods declined to varying degrees. This shows that in weakly supervised fault diagnosis tasks, it is still a challenging problem to ensure the effectiveness of the diagnosis method in cross-domain fault diagnosis tasks. Nevertheless, {\color{black}owing to the improvement of clustering characteristics of low-dimensional features by SSIDEC, and VAT regularization that reduces the overfitting of the diagnostic model, the method proposed in this paper is robust to changes in bearing operating conditions. Over 90\% accuracies are achieved in different cross-domain fault diagnosis tasks.}

The experimental results of tasks C2, C3, and C4 show that in cross-domain fault diagnosis tasks, when the unsupervised data of training and the testing data are under the same operating conditions, the proposed method perform better. Specifically, the diagnostic result of task C2 with different distributions of unsupervised and testing data is lower than that of tasks C3 and C4, and task C3 achieves the best diagnosis results as the distributions of unsupervised and test data are same.

\begin{figure}[!htbp]
	\centering
	
	\subfigure{
		\begin{minipage}[t]{0.4\linewidth}
			\centering
			\includegraphics[width=2.5in]{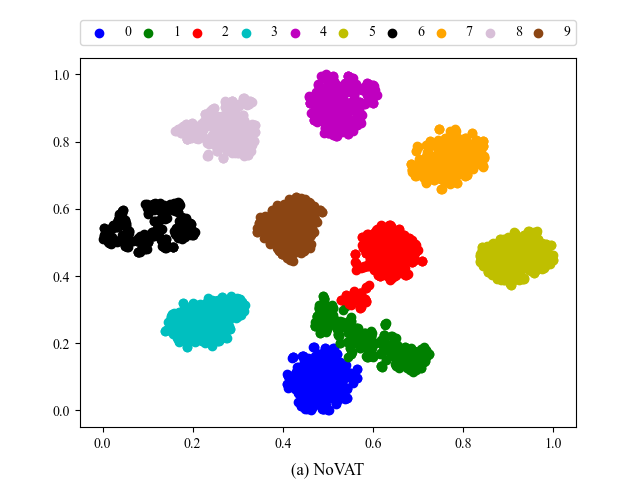}
		\end{minipage}%
	}%
	\subfigure{
		\begin{minipage}[t]{0.4\linewidth}
			\centering
			\includegraphics[width=2.5in]{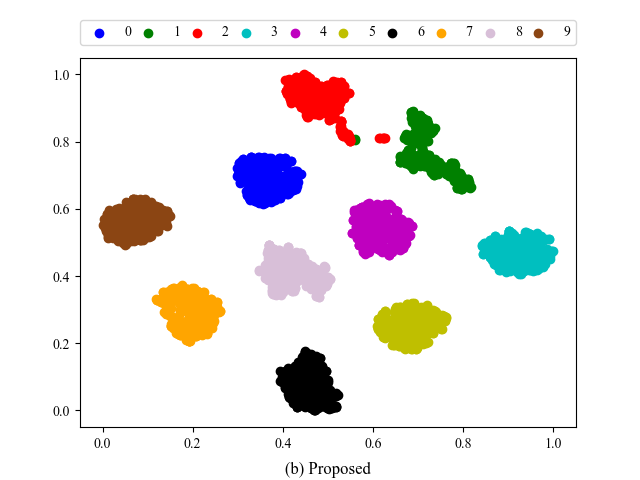}
		\end{minipage}%
	}%
	
	\centering
	\caption{2D data distributions in the task C1 processed by (a)NoVAT method and (b)Proposed method. Ten colors represent ten clusters, respectively. In plot (a), the boundaries of red cluster, green cluster and blue cluster are difficult to divide; in plot (b), the distribution of sample clusters is clearly separated.}
	\label{fig:6}
\end{figure}
\begin{figure}[!htbp]
	\centering
	\subfigure{
		\begin{minipage}[t]{0.4\linewidth}
			\centering
			\includegraphics[width=2.5in]{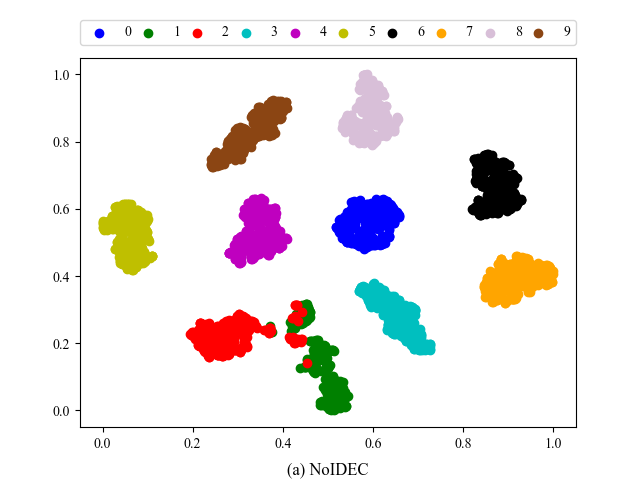}
		\end{minipage}%
	}%
	\subfigure{
		\begin{minipage}[t]{0.4\linewidth}
			\centering
			\includegraphics[width=2.5in]{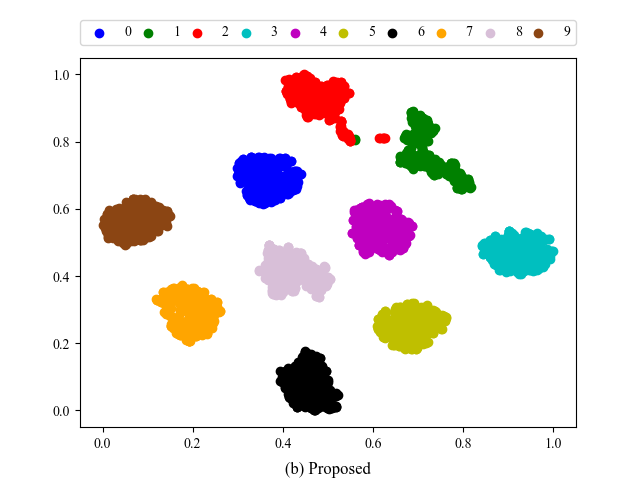}
		\end{minipage}%
	}%
	
	\centering
	\caption{2D data distributions in the task C1 processed by (a)NoIDEC method and (b)Proposed method. Ten colors represent ten clusters, respectively. In plot (a), some of the red clusters and green clusters are mixed together ; in plot (b), the distribution of sample clusters is clearly separated and more concentrated.}
	\label{fig:7}
\end{figure}

Comparing the results of the NoVAT method and the proposed method, it can be seen that the VAT regularization method proposed in this paper can improve the performance of the diagnosis method. Fig. \ref{fig:6} shows the effect of VAT regularization on unsupervised data. Taking task C1 as an example, the accuracy of the proposed method is 1.4\% higher than that of the NoVAT method. This is because VAT regularization can make discrete samples distributed at the edge of the cluster move closer to the cluster center. Therefore, the inter-class distance between different clusters is greater, and the generalization ability of the semi-supervised method {\color{black}is} improved.

The IDEC clustering method {\color{black}significantly} improves the diagnosis result. In task C1, the proposed method is more accurate than the NoIDEC method by 2\%. As shown in Fig. \ref{fig:7}, IDEC training can effectively avoid the aliasing of green and red clusters, which reduces the misclassification between moderate ball failures (Fault 1) and severe ball failures (Fault 2). In addition, IDEC makes the unsupervised data of the same category more densely distributed in the feature space and reduces the intra-class spacing. Since the IDEC improves the {\color{black}clustering accuracy}, the proposed method {\color{black}outperforms other comparison methods} in weakly supervised fault diagnosis tasks.

\begin{table}[htbp]
	\centering
	\caption{The comparison results in the task C5.}
	\label{tab:VAE}       
	\begin{tabular}{lll}
		\hline\noalign{\smallskip}
		VAE(\%) & GAN(\%) & Proposed(\%)  \\
		\noalign{\smallskip}\hline\noalign{\smallskip}
		90.2 & 62.0 & 99.6 \\
		\noalign{\smallskip}\hline
	\end{tabular}
\end{table}
\begin{figure*}[!htbp]
	\centering
	\includegraphics[width=2.5in]{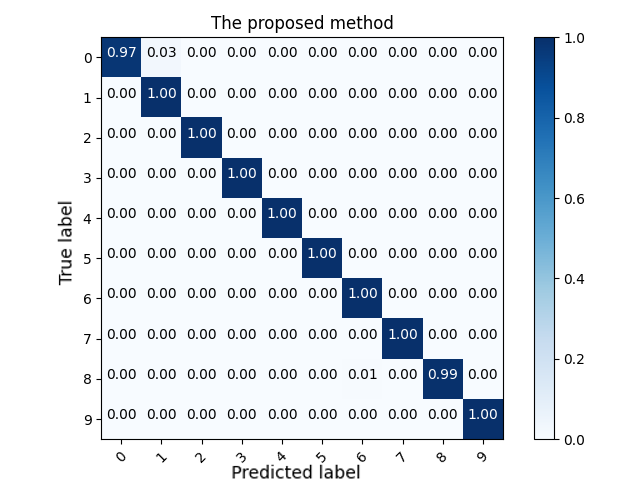}
	\caption{The confusion matrix obtained by the proposed method in task C5. The numerical testing accuracies are presented (\%).}
	\label{cm}       
\end{figure*}

{\color{black}The results of the task C5 are presented in Table \ref{tab:VAE}. The proposed method achieved a better performance than that obtained by the VAE and the GAN. The confusion matrix presented in Fig. \ref{cm} shows the detailed experimental results of the proposed method. It is noted that the lowest accuracy achieved by the proposed method is 97\% for class 0 owing to the general weak features in the ball fault conditions. However, this accuracy is still higher than the average accuracy achieved by VAE and GAN among all classes, which demonstrates the superiority of the proposed method.}

\begin{table}
	\centering
	\caption{Average testing accuracies of the fault diagnosis tasks with weakly supervised data in the MFPT dataset.}
	\label{tab:MFPTresults}       
	\begin{tabular}{llllllll}
		\hline\noalign{\smallskip}
		Task & CNN(\%) & NoIDEC(\%) & NoVAT(\%) & DA(\%) & VAE(\%) & GAN(\%) & Proposed(\%) \\
		\noalign{\smallskip}\hline\noalign{\smallskip}
		M1 & 62.5 & 74.5 & 89.6 & 90.5 & 89.1 & 73.0 & 91.3 \\
		\noalign{\smallskip}\hline
	\end{tabular}
\end{table}

\begin{figure*}[!htbp]
	\centering\includegraphics[width=2.5in]{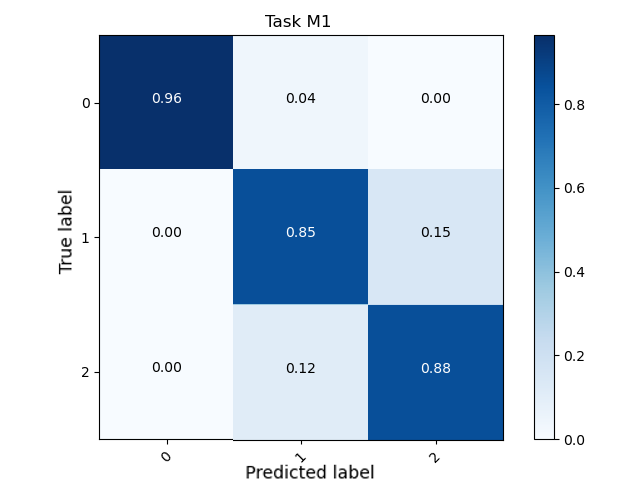}
	\caption{The confusion matrix obtained by the proposed method in the task M1. The numerical testing accuracies are presented (\%).}
	\label{MFPTcm}       
\end{figure*}

{\color{black}The experimental results on the MFPT dataset are presented in Table \ref{tab:MFPTresults}. Since MFPT dataset consists of data under different working conditions, it is more challenging for fault diagnosis. Although the results on this dataset is relative lower than that on task C1 of the CWRU dataset, the proposed method still achieves the best performance among all methods. The accuracy of 91.3 \% achieved by the proposed method is higher than that achieved by DA and VAE. Besides, the proposed method does not require the special preprocessing. Comparing to baselines like CNN, NoIDEC, and NoVAT, the proposed method is competitive in this task. A confusion matric of the proposed method in the task M1 is shown in Fig. \ref{MFPTcm}. Over 80\% testing accuracies can be achieved in every type of fault, which present promising diagnosis accuracies of the proposed method.}

\begin{table}[htbp]
	\centering
	\caption{Diagnostic accuracies achieved by the proposed method after different stages in the task C5.}
	\label{tab:stage}       
	\begin{tabular}{lll}
		\hline\noalign{\smallskip}
		Stage1(\%) & Stage2(\%) & Stage3(\%)\\
		\noalign{\smallskip}\hline\noalign{\smallskip}
		94.03 & 95.48 & 97.6\\
		\noalign{\smallskip}\hline
	\end{tabular}
\end{table}
\begin{figure}[!htbp]
	\centering
	\vspace{-0.4cm}
	\subfigure{
		\begin{minipage}[t]{0.3\linewidth}
			\centering
			\includegraphics[width=2in]{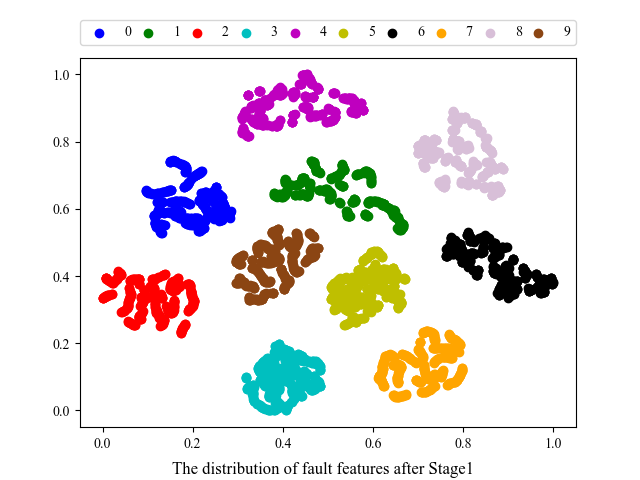}
		\end{minipage}%
	}%
	\subfigure{
		\begin{minipage}[t]{0.3\linewidth}
			\centering
			\includegraphics[width=2in]{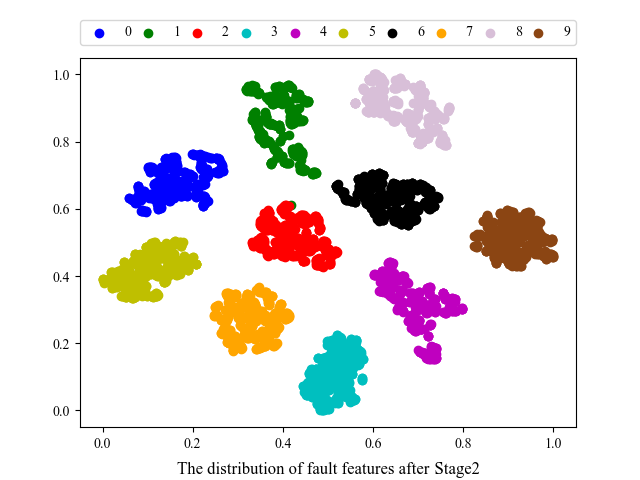}
		\end{minipage}%
	}%
	\subfigure{
		\begin{minipage}[t]{0.3\linewidth}
			\centering
			\includegraphics[width=2in]{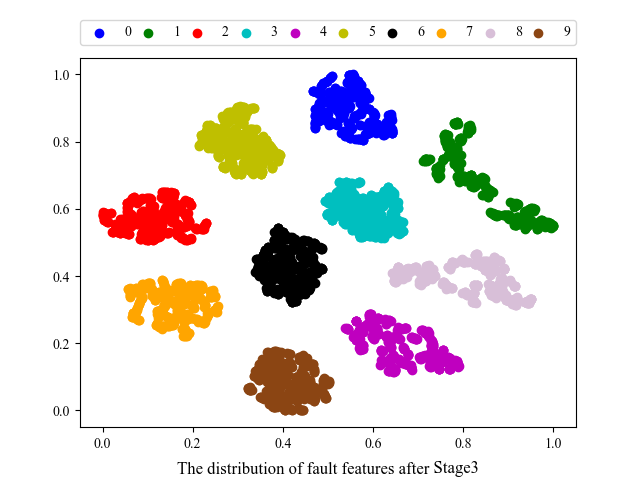}
		\end{minipage}%
	}%
	\centering
	\caption{2D data distributions in the task C5 of features obtained after (a) stage 1, (2) stage 2, and (3) stage 3.}
	\label{disstage}
\end{figure}

\begin{figure}[!htbp]
	\centering
	\vspace{-0.4cm}
	\subfigure{
		\begin{minipage}[t]{0.3\linewidth}
			\centering
			\includegraphics[width=2in]{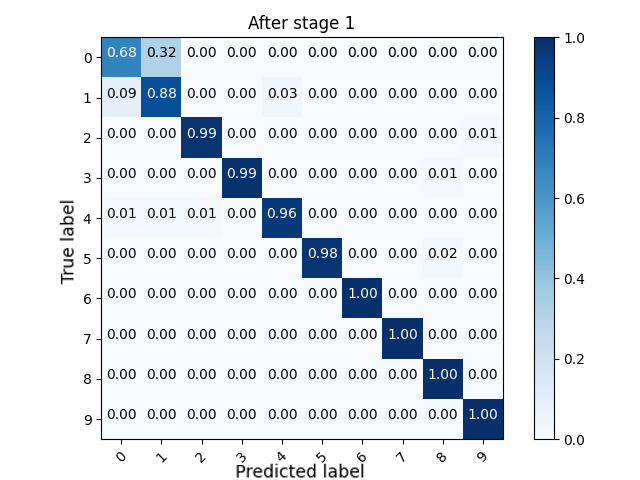}
		\end{minipage}%
	}%
	\subfigure{
		\begin{minipage}[t]{0.3\linewidth}
			\centering
			\includegraphics[width=2in]{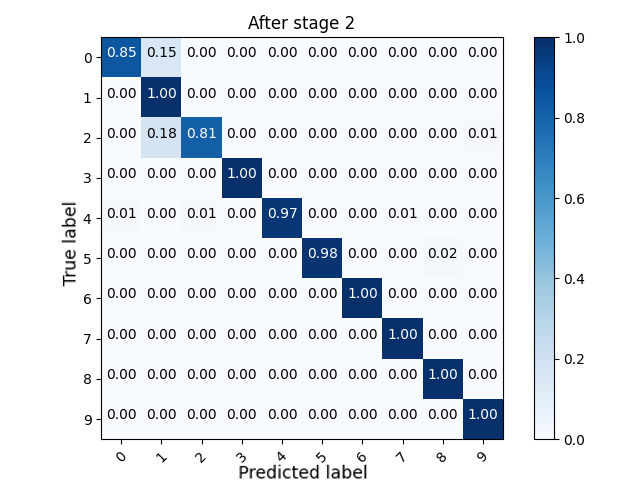}
		\end{minipage}%
	}%
	\subfigure{
		\begin{minipage}[t]{0.3\linewidth}
			\centering
			\includegraphics[width=2in]{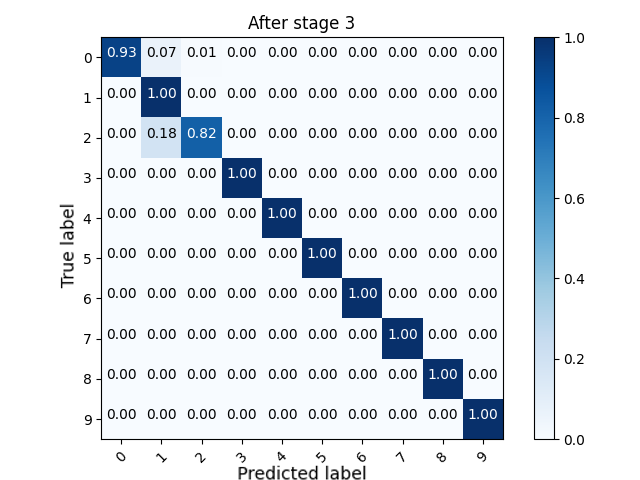}
		\end{minipage}%
	}%
	\centering
	\caption{The confusion matrices in the task C5 after (a) stage 1, (2) stage 2, and (3) stage 3. The numerical testing accuracies are presented (\%).}
	\label{cmstage}
\end{figure}

{\color{black}In tasks with weakly supervised data, the diagnostic process of the proposed method consists of three stages: pre-training, semi-supervised improved deep embedded clustering, and enhanced supervised learning. To show the efficiency of each stage, the diagnostic accuracies, feature distributions, and confusion matrices obtained after different stages are presented in Table \ref{tab:stage}, Fig. \ref{disstage} and Fig. \ref{cmstage}, respectively. It can be noted that, from stage 1 to stage 3, the inter-class distance among different fault types becomes increasingly larger, and the intra-class distance becomes smaller, and thus the generalization and robustness of the proposed method are improved to obtain better accuracies.}

In this section, experimental comparisons of different methods prove the effectiveness and advancement of the proposed method in the task of bearing fault diagnosis under weak supervision. Through the mining of unsupervised data, the diagnostic performance of the semi-supervised learning method is improved when there are limited labeled samples. Meanwhile, the proposed method is robust to changes in working conditions.

\subsection{Performance with unsupervised data}
\label{sec:14}
\subsubsection{Experimental setup}
\label{sec:15}

The method proposed can be directly used to solve the problem of unsupervised fault diagnosis. The clustering results of samples can be obtained through the stage 2. In this study, two experimental tasks, UnsupC1 and UnsupM1, are conducted in the CWRU and MFPT datasets to verify the fault diagnosis performance of the proposed and several advanced methods in unsupervised fault diagnosis. {\color{black}In the task UnsupC1, the number of unsupervised samples for each failure type is $n_{un}=300$ and these samples are under a load of 2 hp. Besides, in the UnsupM1, the total number of the unsupervised samples is 10,808.}

To evaluate the effectiveness of the proposed method, some of the latest clustering algorithms are selected for comparison:

\begin{enumerate}[(1)]
	\item k-means 
	
	The k-means algorithm is a traditional shallow clustering algorithm. In this method, the original frequency-domain signal is directly clustered.
	
	\item AE + k-means
	
	This method combines the deep network and a traditional clustering algorithm. For fairness of comparison, the AE shares the same architecture and parameters with the proposed SCCAE.
	
	\item tSNE + k-means
	
	This is an unsupervised learning method that combines manifold learning and clustering. This method does not require parameter settings and directly clusters the original frequency-domain signals.
	
	\item AE + tSNE + k-means
	
	An auto-encoder+manifold learning+shallow clustering method combines a deep network, manifold learning and traditional clustering. For a fair comparison, the auto-encoder is the same as the skip-connection convolutional auto-encoder in the proposed method.
	
	\item DEC \cite{guo2017improved}
	
	The DEC is a typical deep clustering method. Following the settings in the reference \cite{guo2017improved}, the model is set as a fully-connected multilayer perceptron(MLP) with dimensions of 512-500-500-2000-10.
	
	\item GAN \cite{verstraete2020deep}
	
	{\color{black}This method is based on deep convolutional GANs. Fault features are extracted using the GAN and then clustered by k-means. The network structure and parameters are similar to those in the reference \cite{verstraete2020deep}.}
	
	\item Local Manifold Learning of an Autoencoded Embedding (E2LMC) \cite{An2021Deep}  
	
	{\color{black}E2LMC is reported in the reference \cite{An2021Deep}. The network and parameters are set by following the suggestion of the reference \cite{An2021Deep}.}
	
	\item Deep clustering-based method (DCM) \cite{Li2020Deep}
	
	{\color{black}This network has been proposed to combine deep learning and clustering. The network structure and parameters are similar to those in the reference \cite{Li2020Deep}.}
\end{enumerate}

\subsubsection{Evaluation metrics}
\label{sec:16}
In the study, the clustering accuracy (ACC) [36] and normalized mutual information (NMI) [47] are used to evaluate the effectiveness of the proposed method without supervision. ACC and NMI are popular evaluation indicators in the field of machine learning, which can effectively measure the degree of matching between the clustering results and the actual class labels.

The ACC is used to compare the obtained labels with the actual labels of the data and can be defined as,
\begin{align}
ACC=\mathop{\max}_{map}\frac{\sum_{i=1}^{n_{test}}1\left\{y_i^{test}={map\left(c_i^{assign}\right)}\right\}}{n_{test}}
\end{align}
where $c_i^{assign}$ is the cluster assignment of the $i$-th sample obtained by the proposed method, and $map\left(\right)$ represents all possible one-to-one mappings between the cluster and the real class label.

NMI can defined as,
\begin{align}
{NMI}\left(y_i^{test};c_i^{assign}\right)=\frac{2\times{I\left(y_i^{test};c_i^{assign}\right)}}{H\left(y_i^{test}\right)+H\left(c_i^{assign}\right)}
\end{align}
where function $H\left(\right)$ is used to calculate the entropy value, and $I\left(y_i^{test};c_i^{assign}\right)$ represents the mutual information between $y_i^{test}$ and $c_i^{assign}$.

\subsubsection{Diagnosis results}
\label{sec:17}

\begin{table}[htbp]
	\centering
	\vspace{-0.5cm}
	\caption{Average testing accuracies of the unsupervised learning tasks in the CWRU and MFPT datasets.}
	\label{tab:6}       
	\begin{tabular}{ccccccc}
		\hline\noalign{\smallskip}
		\multirow{2}{*}{Task} & \multicolumn{2}{c}{k-means} & \multicolumn{2}{c}{AE+k-means} & \multicolumn{2}{c}{tSNE+k-means} \\
		& ACC(\%) & NMI(\%) & ACC(\%) & NMI(\%) & ACC(\%) & NMI(\%) \\
		\noalign{\smallskip}\hline\noalign{\smallskip}
		UnsupC1 & 64.9 & 85.4 & 79.8 & 89.6 & 98.9 & 98.8 \\
		UnsupM1 & 53.5 & 35.4 & 82.3 & 65.6 & 88.9 & 78.6 \\
		\noalign{\smallskip}\hline
		\hline\noalign{\smallskip}
		\multirow{2}{*}{Method} & \multicolumn{2}{c}{AE+tSNE+k-means} & \multicolumn{2}{c}{E2LMC} & \multicolumn{2}{c}{DEC}\\
		& ACC(\%) & NMI(\%) & ACC(\%) & NMI(\%) & ACC(\%) & NMI(\%) \\
		\noalign{\smallskip}\hline\noalign{\smallskip}
		UnsupC1 & 98.4 & 98.4 & 98.3 & 97.6 & 82.4 & 87.8 \\
		UnsupM1 & 83.8 & 66.1 & 91.3 & 83.2 & 83.2 & 67.5 \\
		\noalign{\smallskip}\hline
		\hline\noalign{\smallskip}
		\multirow{2}{*}{Method} & \multicolumn{2}{c}{DCM} & \multicolumn{2}{c}{GAN} & \multicolumn{2}{c}{Proposed}\\
		& ACC(\%) & NMI(\%) & ACC(\%) & NMI(\%) & ACC(\%) & NMI(\%) \\
		\noalign{\smallskip}\hline\noalign{\smallskip}
		UnsupC1 & 97.2 & 96.1 & 82.0 & 78.0 & 98.8 & 98.8 \\
		UnsupM1 & 94.3 & 85.2 & 96.0 & 88.0 & 95.6 & 85.9\\
		\noalign{\smallskip}\hline
	\end{tabular}
\end{table}

\begin{figure}[!htbp]
	\centering
	\vspace{-0.5cm}
	\subfigure{
		\begin{minipage}[t]{0.4\linewidth}
			\centering
			\includegraphics[width=2.5in]{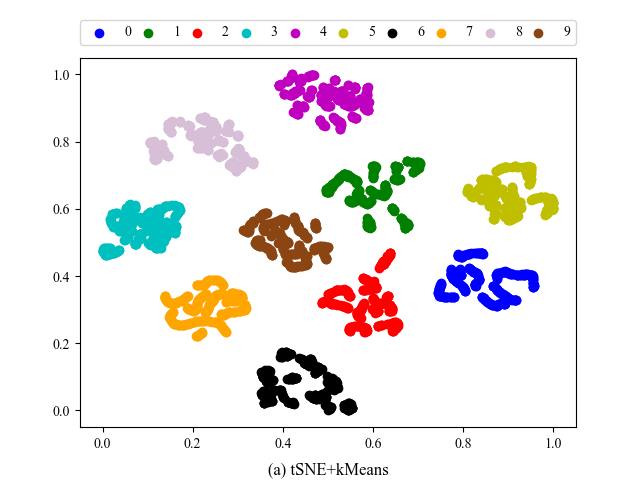}
		\end{minipage}%
	}%
	\subfigure{
		\begin{minipage}[t]{0.4\linewidth}
			\centering
			\includegraphics[width=2.5in]{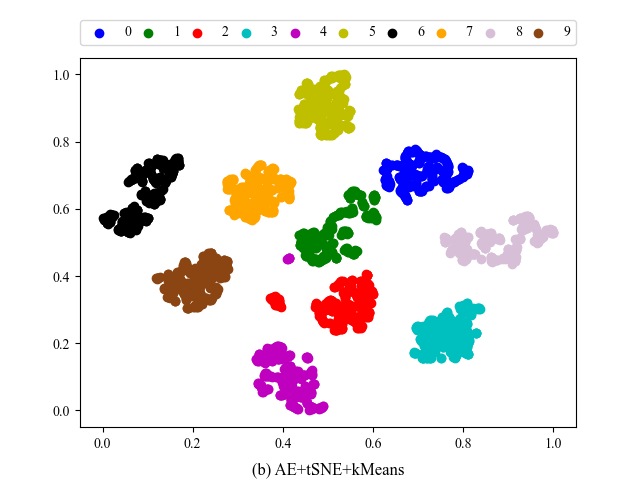}
		\end{minipage}%
	}%
	\quad
	\subfigure{
		\begin{minipage}[t]{0.4\linewidth}
			\centering
			\includegraphics[width=2.5in]{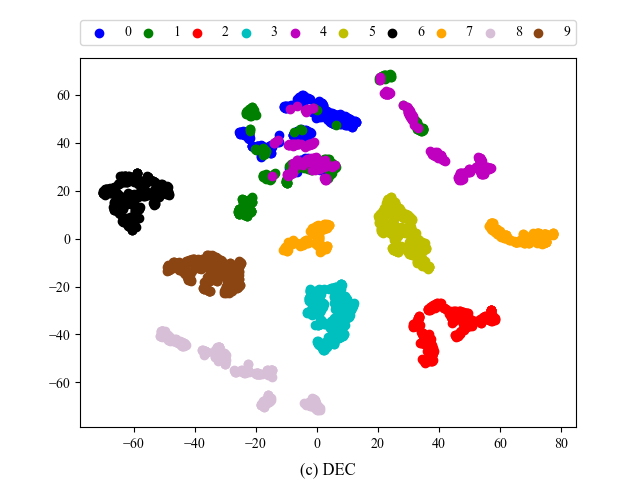}
		\end{minipage}
	}%
	\subfigure{
		\begin{minipage}[t]{0.4\linewidth}
			\centering
			\includegraphics[width=2.5in]{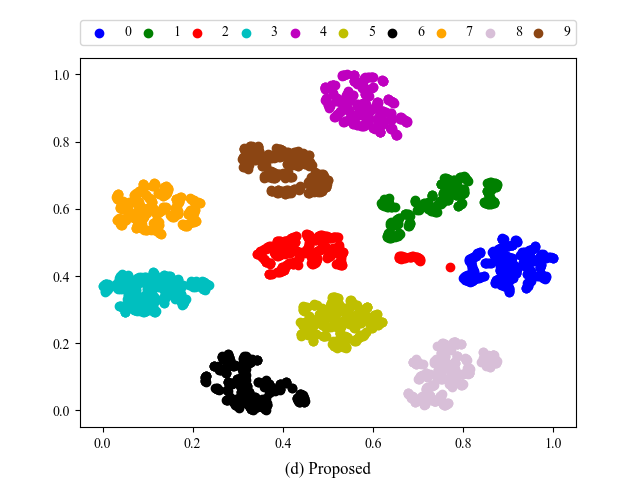}
		\end{minipage}
	}%
	
	\centering
	\caption{2D data distribution in the UnsupC1 processed by (a)tSNE+k-means method, (b)AE+tSNE+k-means method, (c)DEC method and (d)Proposed method. Ten colors represent ten clusters, respectively.}
	\label{fig:8}
\end{figure}

{\color{black}As shown in Table \ref{tab:6}, in the task UnsupC1,} the average results show that the proposed method obtains over 97\% accuracy, which is the highest {\color{black}among all methods}. The result of the AE+k-means method is much higher than that of the k-means method, which proves that the proposed SCCAE is more suitable for clustering. {\color{black}For example, in task UnsupC1,} the visualization of data representations is shown in Fig. \ref{fig:8}. It can be noted that the data representations obtained by the tSNE+k-means method maintain a certain inter-class distance among different clusters but the intra-class distance between the same class of data representation is relatively large. For representations obtained by the AE+tSNE+k-means, although the intra-class distance is smaller than that of the tSNE+k-means method, the inter-class spacing between the blue, green and red samples is too small, which {\color{black}easily leads to} incorrect clustering. The proposed method has better intra-class and inter-class distance than the previous two methods, {\color{black}and thus} the clustering results are the best. The original DEC method can also be used for unsupervised bearing-fault diagnosis tasks, but the preformance is not competitive as this method does not introduce the SCCAE or consider the reconstruction loss to maintain the local structure of the data. {\color{black}For the MFPT dataset, similar results are achieved as well. Only the GAN-based method achieves a good performance as well as the proposed method. However, it does not perform well in the UnsupC1. In sum, the promising diagnostic performance of the proposed method can be validated from the results in the UnsupC1 and UnsupM1 tasks.}

In this section, by comparing the experimental results of different methods in unsupervised bearing fault diagnosis tasks, it is proved that the proposed method can be extended to unsupervised bearing fault diagnosis tasks and obtain high-quality clustering results.

\subsection{Model performance analysis}
\label{sec:18}
In this section, the effects of the amount of supervised data $n_{sp}$, the dimension of the low-dimensional feature vector $N_{rep}$ and the clustering loss coefficient $\gamma_c$ on the proposed method is studied {\color{black}in the CWRU dataset.} 

\begin{figure}[!htbp]
	\centering
	\includegraphics[width=0.8\textwidth]{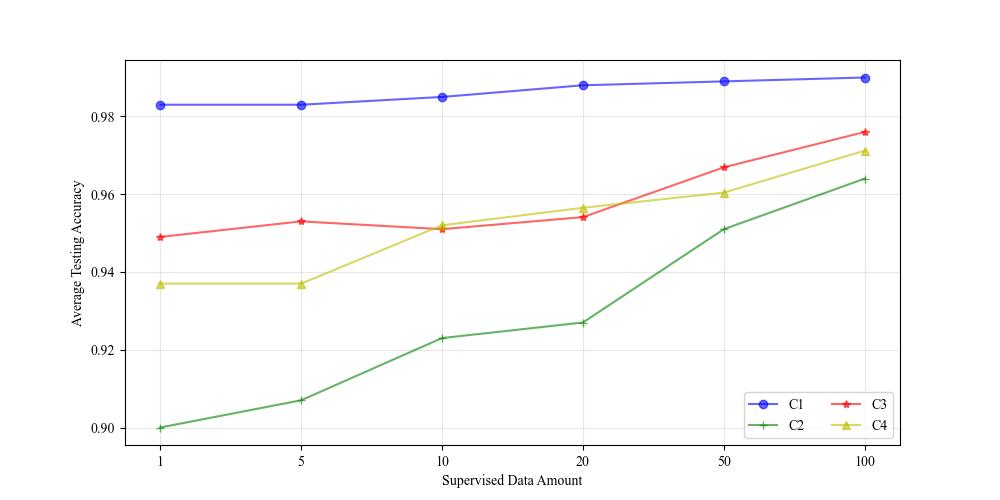}
	\caption{Average testing accuracy of different tasks with different amount of unsupervised data in the CWRU dataset.}
	\label{fig:9}       
\end{figure}

First, the influence of $n_{sp}$ on {\color{black}testing} accuracy is studied. As shown in Fig. \ref{fig:9}, in the same task, the greater the number of labeled samples, the higher is the test accuracy achieved. This shows that the proposed method has achieved good performance in the case of scarce labeled samples, and more labeled samples can further improve the generalization ability of the method.

\begin{figure}[!htbp]
	\centering
	\includegraphics[width=0.8\textwidth]{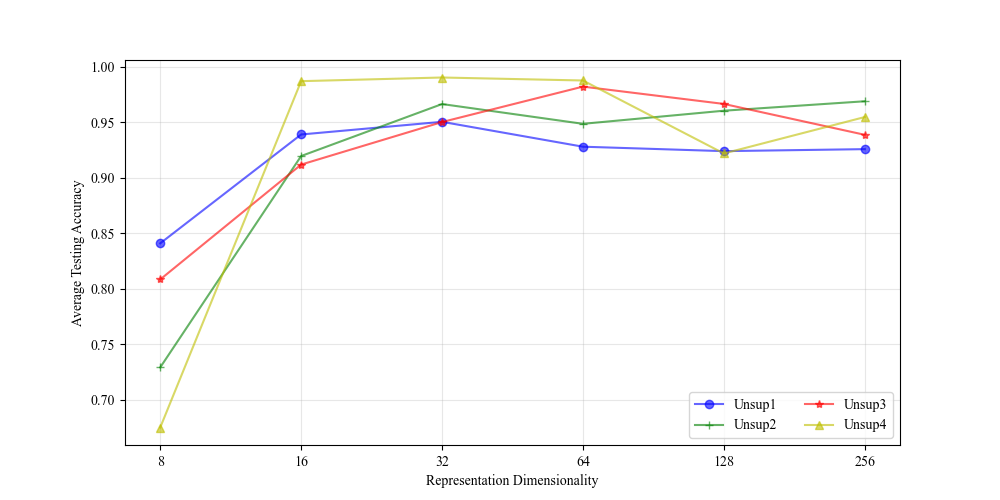}
	\caption{Average testing accuracy of different tasks with different low-dimensional representation dimensionality.}
	\label{fig:10}       
\end{figure}
\begin{figure}[!htbp]
	\centering
	\vspace{-0.5cm}
	\includegraphics[width=0.8\textwidth]{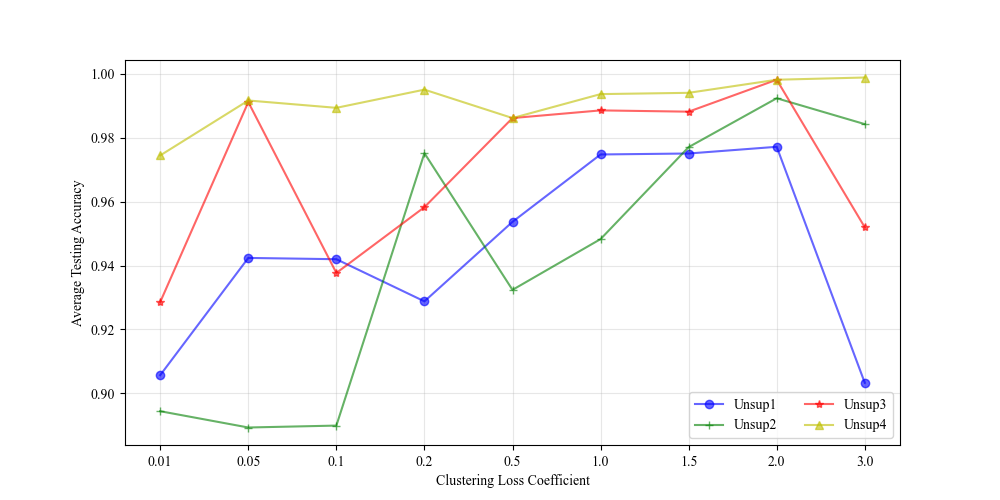}
	\caption{Average testing accuracy of different tasks with different clustering loss coefficient in the CWRU dataset.}
	\label{fig:11}       
\end{figure}

Subsequently, the influence of $N_{rep}$ on the accuracy of the test is studied {\color{black}in the CWRU dataset}. As shown in Fig. \ref{fig:10}, in all tasks, the proposed method performs worst with $N_{rep}=8$, which illustrates that the feature with too low dimensions cannot contain {\color{black}sufficient fault information.} When dimensions are greater than 16 and less than 64, the change in dimension has little effect on the test accuracy. When the dimensionality is greater than 64, the test accuracy has a small decrease, which shows that the features of larger dimensionality increase the difficulty of training the model. Therefore, {\color{black}a moderate number of dimensions of low-dimensional representation should be determined in the study.}

IDEC learns clustering-friendly representations by reducing the clustering loss. The influence of $\gamma_c$ on the test accuracy is shown in Fig. \ref{fig:11}. It can be seen that in different tasks, as the clustering loss coefficient increases, the test accuracy increases to varying degrees. The larger the clustering coefficient, the greater the contribution of the clustering loss to the training of the model, {\color{black}and thus} the learned low-dimensional representation is more suitable for clustering. However, excessive clustering loss distorts the feature space and destroys the local structure of the data features. Therefore, within a certain range, the increase helps to improve the test accuracy. Beyond this range, the increase actually reduces the test accuracy.

\section{Conclusions}
\label{sec:19}
In this study, a MS-SSIDEC method is proposed for bearing fault diagnosis {\color{black}under the situation of insufficient labeled data}. This method combines the advantages of semi-supervised learning with deep clustering. {\color{black}A three-stage training strategy is adopted, that is pre-training, semi-supervised improved deep embedded clustering, and enhanced supervised learning.} The SCCAE in the first stage can automatically learn low-dimensional features containing more meaningful fault information. The SSIDEC method proposed in the second stage can make the learned representations more clustering-friendly and obtain high-quality clustering assignments of unsupervised data. The pseudo-labeled samples can be used to enrich the limited labeled dataset. To overcome the overfitting, VAT is introduced as a regularization term. In the third stage, the augmented dataset is used to train a bearing-fault discriminative model with a strong generalization ability. {\color{black}The experimental results on the CWRU and MFPT bearing datasets prove that this method can solve the semi-supervised fault diagnosis problem in the case of limited labeled samples,} and the method can also be effectively extended to unsupervised fault diagnosis tasks. This method provides promising research direction for solving the problem of fault diagnosis under the situation of insufficient labeled samples.
Future research will focus on two aspects: (1) To explore deep clustering methods that can be applied in the case of unknown number of clusters. (2) To improve the method's ability to diagnose bearing faults under variable operating conditions so that it can be applied in more real industrial scenarios.

\bibliography{Manuscript}

\end{document}